\documentclass{article}

\PassOptionsToPackage{numbers, compress}{natbib}

\usepackage[preprint]{neurips_2023}

\usepackage[utf8]{inputenc} 
\usepackage[T1]{fontenc}    
\usepackage{hyperref}       
\usepackage{url}            
\usepackage{booktabs}       
\usepackage{amsfonts}       
\usepackage{nicefrac}       
\usepackage{microtype}      
\usepackage{xcolor}         

\usepackage{amsmath}
\usepackage{graphicx}
\usepackage{mathtools}
\usepackage{adjustbox}

\newcommand{\R}{\mathbb{R}}
\newcommand{\bI}{\mathbf{I}}

\newcommand{\hby}{\hat{\mathbf{y}}}
\newcommand{\bG}{\mathbf{G}}
\newcommand{\bR}{\mathbf{R}}
\newcommand{\bA}{\mathbf{A}}
\newcommand{\bH}{\mathbf{H}}

\DeclareMathOperator{\ReLU}{ReLU}

\def\secref#1{Section~\ref{#1}}
\def\figref#1{Figure~\ref{#1}}
\def\tabref#1{Table~\ref{#1}}
\def\eqref#1{Eq.~\ref{#1}}

\def\lstref#1{Listing~\ref{#1}}

\usepackage{derivative}
\usepackage{subcaption}  
\usepackage{bbding}  
\usepackage{lipsum}
\usepackage{multirow}
\usepackage{listings}
\usepackage{xcolor}

\definecolor{codegreen}{rgb}{0,0.6,0}
\definecolor{codegray}{rgb}{0.5,0.5,0.5}
\definecolor{codepurple}{rgb}{0.58,0,0.82}
\definecolor{backcolour}{rgb}{0.95,0.95,0.92}

\lstdefinestyle{mystyle}{
    backgroundcolor=\color{backcolour},   
    commentstyle=\color{codegreen},
    keywordstyle=\color{magenta},
    numberstyle=\tiny\color{codegray},
    stringstyle=\color{codepurple},
    basicstyle=\ttfamily\footnotesize,
    breakatwhitespace=false,         
    breaklines=true,                 
    captionpos=b,                    
    keepspaces=true,                 
    numbers=left,                    
    numbersep=5pt,                  
    showspaces=false,                
    showstringspaces=false,
    showtabs=false,                  
    tabsize=2
}

\title{Resolution-Aware Design of Atrous Rates for Semantic Segmentation Networks}

\author{%
    Bum Jun Kim\\
    POSTECH\\
  \texttt{kmbmjn@postech.edu} \\
  \And
  Hyeyeon Choi\\
    POSTECH\\
  \texttt{hyeyeon@postech.edu} \\
  \And
    Hyeonah Jang\\
    POSTECH\\
  \texttt{hajang@postech.edu} \\
  \And
    Sang Woo Kim\\
    POSTECH\\
  \texttt{swkim@postech.edu} \\
}

\begin{document}

\maketitle

\begin{abstract}
	DeepLab is a widely used deep neural network for semantic segmentation, whose success is attributed to its parallel architecture called atrous spatial pyramid pooling (ASPP). ASPP uses multiple atrous convolutions with different atrous rates to extract both local and global information. However, fixed values of atrous rates are used for the ASPP module, which restricts the size of its field of view. In principle, atrous rate should be a hyperparameter to change the field of view size according to the target task or dataset. However, the manipulation of atrous rate is not governed by any guidelines. This study proposes practical guidelines for obtaining an optimal atrous rate. First, an effective receptive field for semantic segmentation is introduced to analyze the inner behavior of segmentation networks. We observed that the use of ASPP module yielded a specific pattern in the effective receptive field, which was traced to reveal the module's underlying mechanism. Accordingly, we derive practical guidelines for obtaining the optimal atrous rate, which should be controlled based on the size of input image. Compared to other values, using the optimal atrous rate consistently improved the segmentation results across multiple datasets, including the STARE, CHASE\_DB1, HRF, Cityscapes, and iSAID datasets.
\end{abstract}

\section{Introduction}
\label{sec:intro}
Semantic segmentation refers to the task of generating a semantic mask that classifies each pixel in an image into a specific category \citep{DBLP:journals/tog/AksoyOPPM18,DBLP:journals/tog/HuangFL14,DBLP:journals/tog/XiaoFZLQ09}. Semantic segmentation is one of the most representative tasks in the field of computer vision and is crucial for understanding scenes in indoor and outdoor environments. The recent success of deep neural networks has been incorporated into semantic segmentation using an encoder-decoder architecture, thereby enabling high performance \citep{DBLP:journals/tog/ZhuAFW21,DBLP:journals/tog/SchneiderT16}.

However, one challenge in semantic segmentation is detecting objects of varying sizes. Using the cascade architecture of deep neural networks leads to single-level image understanding, which complicates the detection of small and large objects within an image. To facilitate image understanding with multi-level features, modern segmentation networks employ a parallel architecture called atrous spatial pyramid pooling (ASPP) atop an encoder, which enables the extraction of both local and global information from the encoded features. Since its introduction in DeepLabV2 \citep{DBLP:journals/pami/ChenPKMY18}, the ASPP module has yielded successful results in semantic segmentation and has been applied to other dense prediction tasks, including instance \citep{DBLP:conf/cvpr/ChenPWXLSF0SOLL19} and panoptic segmentations \citep{DBLP:journals/corr/abs-2011-11675,DBLP:conf/cvpr/ChengCZ0HAC20}, monocular depth estimation \citep{DBLP:journals/corr/abs-1907-10326,DBLP:conf/cvpr/FuGWBT18}, domain adaptation \citep{DBLP:conf/mm/GaoZZT21,DBLP:conf/cvpr/GuoYLY21,DBLP:conf/ijcnn/MarsdenBDY22,DBLP:journals/tip/ZhengY22}, and image matting \citep{DBLP:conf/iccv/0003DSX19}.

\begin{figure}[t!]
	\centering
	\includegraphics[width=0.99\linewidth]{"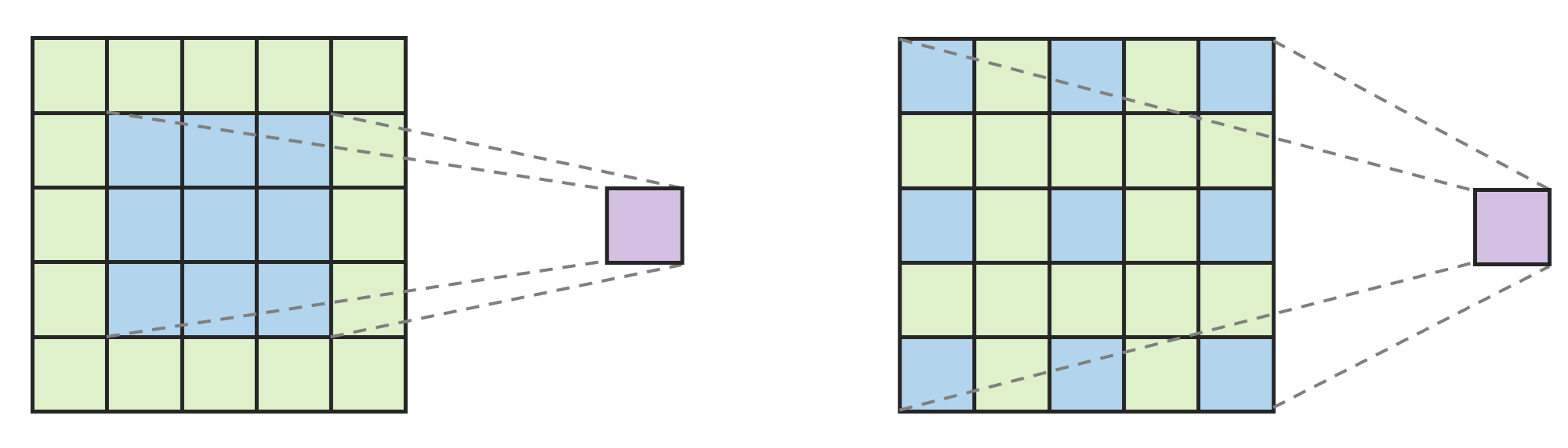"}
	\caption{$3 \times 3$ vanilla convolution uses nine adjoint features (shown in blue on the left), whereas $3 \times 3$ atrous convolution with an atrous rate of 2 uses nine distant features with one feature in between (shown in blue on the right).}
	\label{fig:atrous}
\end{figure}

The aim of ASPP module is to enlarge the field of view (FOV) of the segmentation network using atrous convolution \citep{DBLP:journals/pami/ChenPKMY18}, also known as dilated convolution \citep{DBLP:conf/cvpr/YuKF17,DBLP:journals/corr/YuK15}. In contrast to vanilla convolution, atrous convolution uses an atrous rate, which generates an empty space between each element of the convolutional kernel (\figref{fig:atrous}). After applying multiple atrous convolutions with various atrous rates, the ASPP module combines their outputs, which enables it to detect objects of varied sizes and capture an understanding of the global context of the image, such as the overall layout and relationships between objects.

The FOV size of the segmentation network is determined by the atrous rates of ASPP module and spatial scale of the encoded feature. To date, the common choice for atrous rates of the ASPP module has been $\{6, 12, 18\}$ for three atrous convolutions or their doubled values when using a lower downsampling rate. However, in this study, we demonstrate that the existing rule for determining atrous rates yields a fixed FOV size, which poses limitations on its potential benefits by using the correct size. The FOV size should be adjusted by considering the properties of the target task or dataset to obtain more effective behavior of the ASPP module.

To achieve this, the detailed mechanisms of the ASPP module are investigated. Despite their prevalence, the internal behavior of ASPP module in deep neural networks has rarely been discussed in semantic segmentation. To this end, an effective receptive field (ERF) is introduced for semantic segmentation that visualizes a segmentation network. We discover that the ERF of segmentation network exhibits a specific pattern owing to the architectural properties of ASPP module. Based on this pattern, we are going to explain the mechanism of ASPP module. By analyzing the ASPP module, its FOV size is determined and guidelines are proposed for controlling the FOV size by choosing valid values of atrous rates. Finally, the performances of the proposed atrous rates were compared with other values on various semantic segmentation datasets.

\section{Background}
\label{sec:back}

\subsection{ERF for Semantic Segmentation Networks}
Let $\bI \in \R^{H \times W \times C}$ be an input image for a semantic segmentation model, where $(H, W)$ is the resolution or size of the image and $C$ represents the number of channels. The objective of semantic segmentation is to generate a semantic mask $\hby \in \R^{H \times W \times N_c}$ that classifies each pixel in an image $\bI$ into one of the $N_c$ categories. A deep neural network is used as the semantic segmentation model that outputs $\hby$ from the input image $\bI$. Subsequently, the relationship between $\hby$ and $\bI$ is represented by a differentiable function involving various operations including convolution, batch normalization, and ReLU.

ERF has been widely used for image classification \citep{DBLP:conf/nips/LuoLUZ16,DBLP:journals/prl/KimCJLJK23}; however, ERF for semantic segmentation has been rarely discussed. Therefore, we first formulate an ERF for semantic segmentation. The objective is to analyze an input pixel-level area that influences a pixel unit in the segmentation output. The difference between ERFs for image classification and semantic segmentation depends on the choice of the target unit in the output. Herein, the pixel located at the central coordinate $(C_h, C_w)$ is selected to examine the spatial relationship between the central output unit and input image. The central output unit is defined as $Y \coloneqq \sum_{k=1}^{N_c}{\hat{y}_{C_h,C_w,k}} \in \R$, which is aggregated across all categories. Because a gradient indicates the sensitivity of the pixel \citep{DBLP:journals/corr/SimonyanVZ13}, the contribution of each pixel in $\bI$ to $Y$ can be investigated using the gradient $\pdv{Y}{\bI} \in \R^{H \times W \times C}$. After summing the gradient over the image channels, $\bG \coloneqq \sum_{c=1}^{C} \pdv{Y}{\bI} \in \R^{H \times W}$ is obtained, which represents the influence of each pixel in the input image on the central output unit. However, the gradient obtained from a single image is sparse and highly dependent on the properties of the input image. To examine the general behavior of the segmentation network, $\bG$ is aggregated across a large image dataset $S$. Additionally, $\ReLU$ is used to filter the negative and accumulate the positive importances \citep{DBLP:journals/ijcv/SelvarajuCDVPB20}. Therefore, the ERF $\bR \coloneqq \sum_{I \in S} \ReLU(\bG) \in \R^{H \times W}$ is obtained, which illustrates the general contribution of each pixel to the central output unit. Note that ERF is defined as the relationship between the input and output of the segmentation network, which includes the properties of encoder and decoder. In summary, the ERF is obtained as follows:
\begin{align}
	\bR = \sum_{I \in S} \ReLU\left(\sum_{c=1}^{C} \pdv{(\sum_{k=1}^{N_c}{\hat{y}_{C_h,C_w,k}})}{\bI}\right).
\end{align}

\subsection{ASPP Module}
\label{sec:aspp}

\begin{figure}[t!]
	\begin{lstlisting}[language=Python, label=lst:aspp, caption=One example for the implementation of the ASPP module of DeepLabV3+ \citep{ssp} in PyTorch \citep{DBLP:conf/nips/PaszkeGMLBCKLGA19}.]
"""
One example for the implementation of the ASPP module of DeepLabV3+ in PyTorch.
Reference: https://github.com/yassouali/pytorch-segmentation/blob/master/models/deeplabv3_plus.py
"""
class ASSP(nn.Module):
    def __init__(self, in_channels, output_stride):
        super(ASSP, self).__init__()

        assert output_stride in [8, 16], 'Only output strides of 8 or 16 are suported'
        if output_stride == 16: dilations = [1, 6, 12, 18]
        elif output_stride == 8: dilations = [1, 12, 24, 36]

        self.aspp1 = assp_branch(in_channels, 256, 1, dilation=dilations[0])
        self.aspp2 = assp_branch(in_channels, 256, 3, dilation=dilations[1])
        self.aspp3 = assp_branch(in_channels, 256, 3, dilation=dilations[2])
        self.aspp4 = assp_branch(in_channels, 256, 3, dilation=dilations[3])

        # ...
\end{lstlisting}
\end{figure}

For an input image $\bI \in \R^{H \times W \times C}$, the encoder of a segmentation network produces a feature map $\bH \in \R^{(H/s) \times (W/s) \times M}$, where $M$ represents the number of channels for the feature map and $s$ denotes the \textit{output stride}, indicating the downsampling ratio up to the encoder output. The output stride value is generally selected as $s=8$ or $s=16$. A study of DeepLabV3 \citep{DBLP:journals/corr/ChenPSA17} observed that using $s=8$ resulted in improved accuracy with additional use of computational resources, whereas $s=16$ provided reasonable performance and $s \geq 32$ caused performance degradation. This is because using higher downsampling rates eliminates fine details in images, thereby decreasing the accuracy of dense prediction tasks, such as semantic segmentation.

After obtaining the encoder output $\bH$, DeepLabV3 and its variants apply the ASPP module containing the following five branches: one $1 \times 1$ convolution, one image pooling, and three $3 \times 3$ atrous convolutions with atrous rates of $\{r, 2r, 3r\}$. The results from these branches are concatenated and subsequently merged with additional convolutions to produce the ASPP output $\bA \in \R^{(H/s) \times (W/s) \times N_c}$. Finally, bilinear upsampling is applied to $\bA$ for obtaining the final dense segmentation $\hby \in \R^{H \times W \times N_c}$.

The three atrous rates $\{r, 2r, 3r\}$ of the ASPP module are determined by the base atrous rate $r$. To choose the base atrous rate, DeepLabV3 \citep{DBLP:journals/corr/ChenPSA17} and DeepLabV3+ \citep{DBLP:conf/eccv/ChenZPSA18} use the following rule:
\begin{align}
	r = \begin{cases}
		    6  & \text{if $s = 16$,} \\
		    12 & \text{if $s = 8$.}
	    \end{cases} \label{eq:bar}
\end{align}
This rule is widely deployed in numerous semantic segmentation libraries and codes (\lstref{lst:aspp}). However, the two cases are later demonstrated to be equivalent because enlarging FOV by increasing atrous rates has the same effect as decreasing the downsampling rate (\secref{sec:howaspp}). Thus, assuming $s=16$, DeepLabV3 and DeepLabV3+ employ an ASPP module with atrous rates of $\{6, 12, 18\}$.

The base atrous rate $r=6$ originates from an earlier version. In an ablation study of DeepLabV2 \citep{DBLP:journals/pami/ChenPKMY18}, ASPP-S with four branches of atrous rates $\{2, 4, 8, 12\}$ and ASPP-L with $\{6, 12, 18, 24\}$ were compared, concluding that the latter performed better on the PASCAL VOC 2012 dataset \citep{DBLP:journals/ijcv/EveringhamEGWWZ15}. In DeepLabV3, the fourth atrous branch was removed using the three atrous rates, $\{6, 12, 18\}$. Since then, the ASPP module has been widely employed with $r=6$ as the default value in semantic segmentation.

A default value of $r=6$ can be used; however, it is not guaranteed to be optimal. In principle, the base atrous rate $r$ should be a hyperparameter to change the FOV size of the segmentation network based on the target task or dataset. The identification of the optimal value of base atrous rate can improve the performance of segmentation networks compared to a suboptimal one. However, few studies have attempted to determine the optimal atrous rate, and the default value of $r=6$ has been simply used without considering the specific properties of target task or dataset. References \citep{DBLP:conf/nips/ChenCZPZSAS18} and \citep{DBLP:conf/cvpr/LiuCSAHY019} exploited a neural architecture search to automatically discover an improved architecture for the ASPP module; however, these approaches do not consider dataset dependency because the search is based on a single dataset, and do not provide a logical understanding of the ASPP module.

We claim that the mechanism of the ASPP module inside deep neural networks has not been thoroughly understood. Owing to this difficulty, there are currently no guidelines on exactly how much we should control the atrous rates to enlarge or shrink the FOV. Furthermore, because the default value $r=6$ was obtained from the ablation study using a single dataset of the PASCAL VOC 2012, its validity should be verified across multiple datasets. In practical scenarios requiring a smaller FOV, using a base atrous rate other than $r=6$ can be beneficial. Conversely, for larger image sizes, designing a larger FOV using a different base atrous rate can be advantageous. To this end, this study aims to establish practical guidelines for obtaining an optimal atrous rate.

\section{Understanding the ASPP Module}
\label{sec:understanding}

\subsection{Analysis of ERF for Semantic Segmentation Networks}
\label{sec:analsis}

\begin{figure}[t!]
	\centering
	\begin{subfigure}[b]{0.241\linewidth}
		\centering
		\includegraphics[width=\linewidth]{"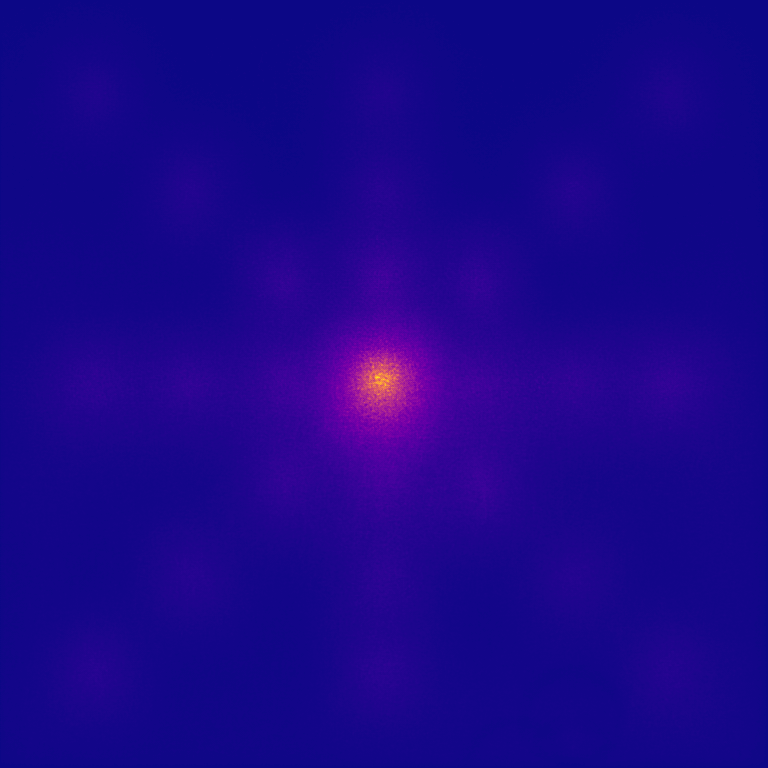"}
		\caption{DeepLabV3 with R-18, $s=8$}
	\end{subfigure}
	\hfill
	\begin{subfigure}[b]{0.241\linewidth}
		\centering
		\includegraphics[width=\linewidth]{"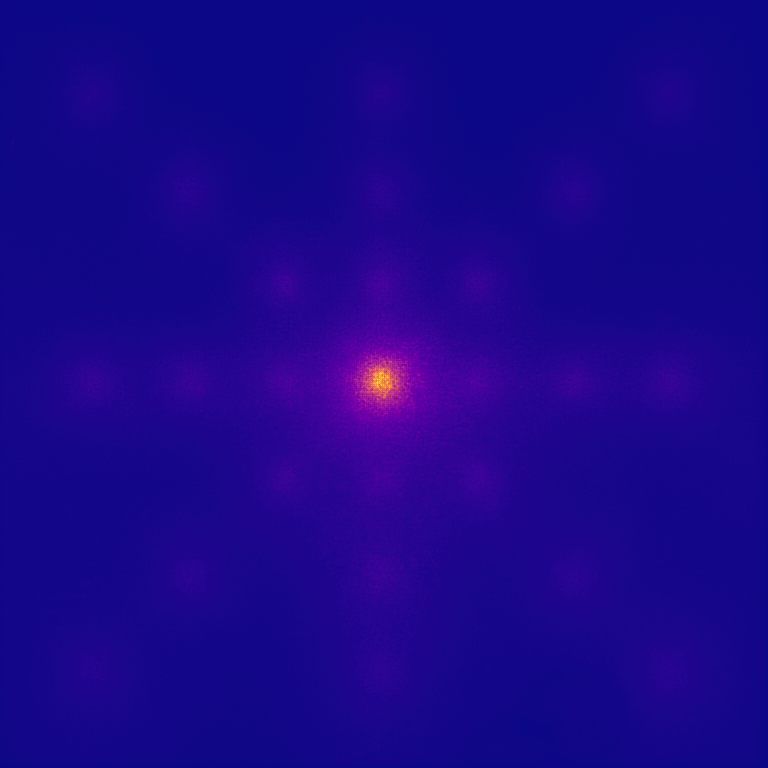"}
		\caption{DeepLabV3 with R-50, $s=8$}
	\end{subfigure}
	\hfill
	\begin{subfigure}[b]{0.241\linewidth}
		\centering
		\includegraphics[width=\linewidth]{"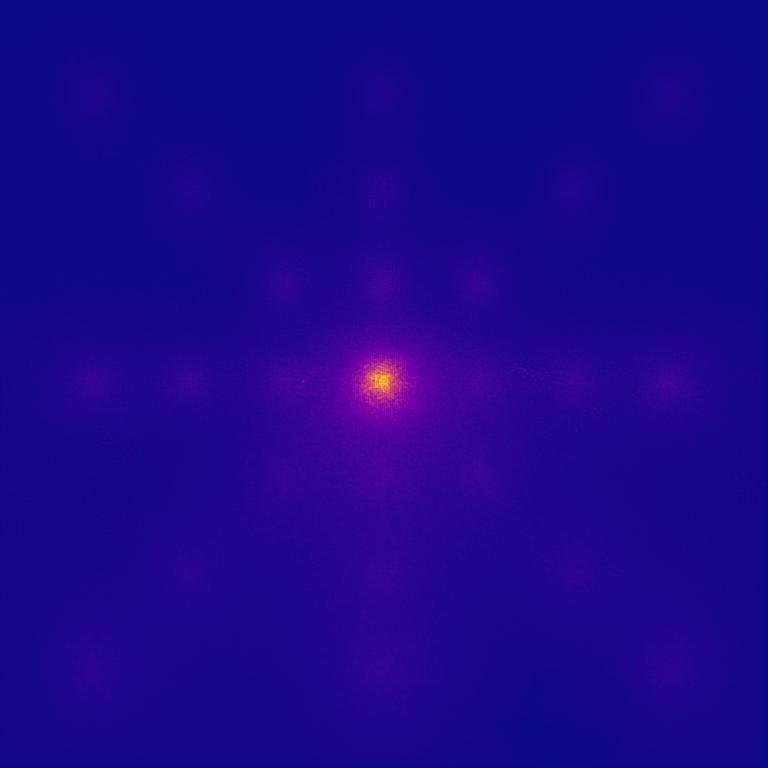"}
		\caption{DeepLabV3 with R-101, $s=8$}
	\end{subfigure}
	\hfill
	\begin{subfigure}[b]{0.241\linewidth}
		\centering
		\includegraphics[width=\linewidth]{"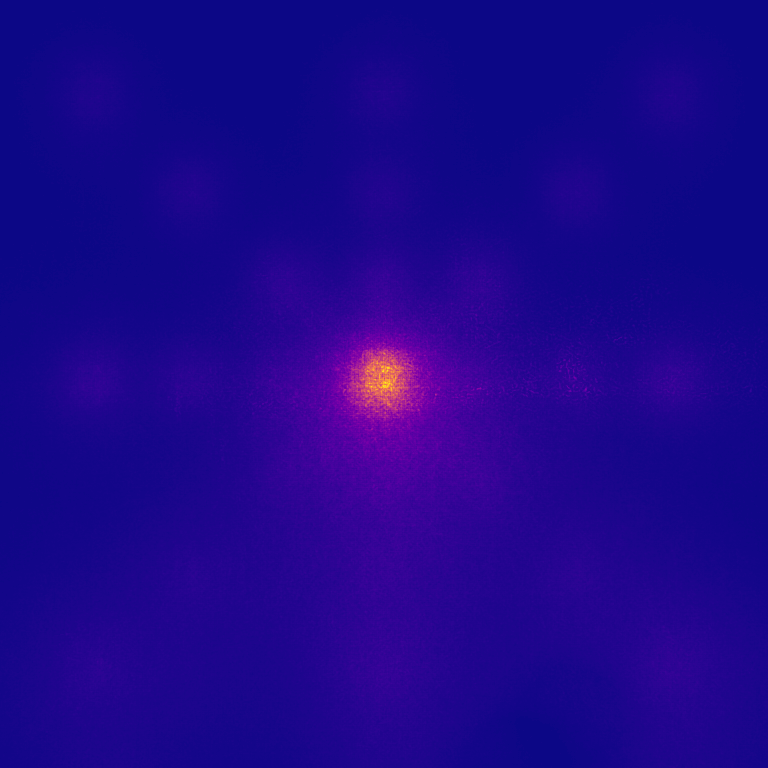"}
		\caption{DeepLabV3 with R-101, $s=16$}
	\end{subfigure}
	\caption{ERFs of DeepLabV3 for the Cityscapes dataset in $768 \times 768$ input image. ``R'' indicates ResNet. We observed that all ERFs of DeepLabV3 exhibited an \EightStarTaper-shaped pattern, which is referred to as the \textit{star pattern}. Because printed figures can be seen improperly, we highly encourage viewing all images electronically with zoom. See supplementary materials to review the raw image files.}
	\label{fig:d3}
\end{figure}

To investigate the inner mechanism of ASPP module, we begin with empirical observations of the segmentation networks. To this end, the ERFs of DeepLabV3 and DeepLabV3+ were analyzed under several conditions, including the used datasets, types of backbones, and choices of output strides. \figref{fig:d3} illustrates the ERFs of DeepLabV3 using the Cityscapes dataset \citep{DBLP:conf/cvpr/CordtsORREBFRS16} with an input size of $768 \times 768$. Note that, for the three backbones of ResNet-\{18, 50, 101\} \citep{DBLP:conf/cvpr/HeZRS16}, the ERFs of DeepLabV3 exhibited an \EightStarTaper-shaped pattern, which we refer to as the \textit{star pattern}. Although the star patterns overlook the less-used areas, their broad coverage enables the segmentation network to capture the global context of an input image. Additionally, when measuring the end-to-end distance, a star pattern of the same size was observed for the two output strides $s \in \{8, 16\}$.

\begin{figure}[t!]
	\centering
	\begin{subfigure}[b]{0.241\linewidth}
		\centering
		\includegraphics[width=\linewidth]{"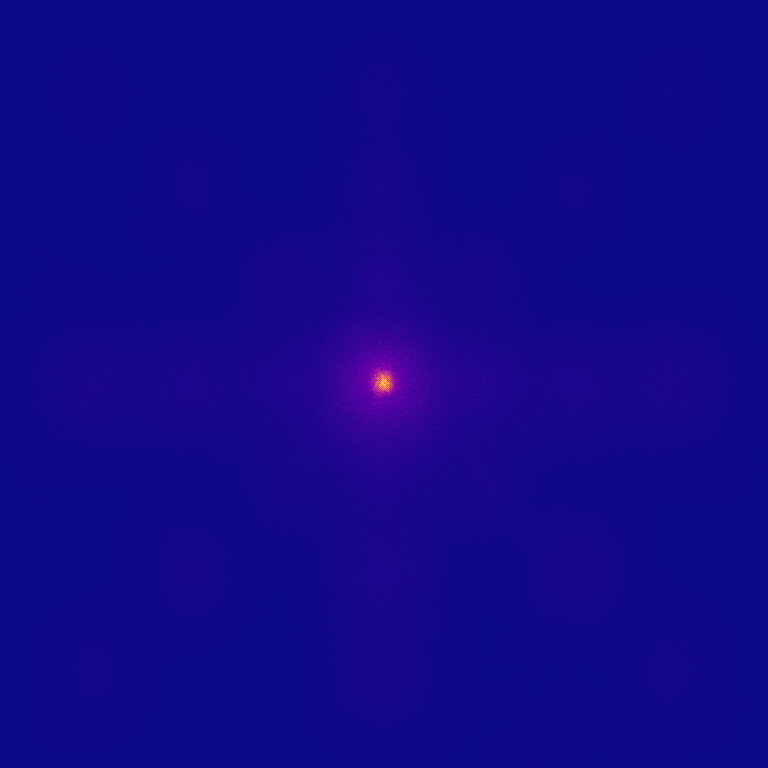"}
		\caption{DeepLabV3+ with R-18, $s=8$}
	\end{subfigure}
	\hfill
	\begin{subfigure}[b]{0.241\linewidth}
		\centering
		\includegraphics[width=\linewidth]{"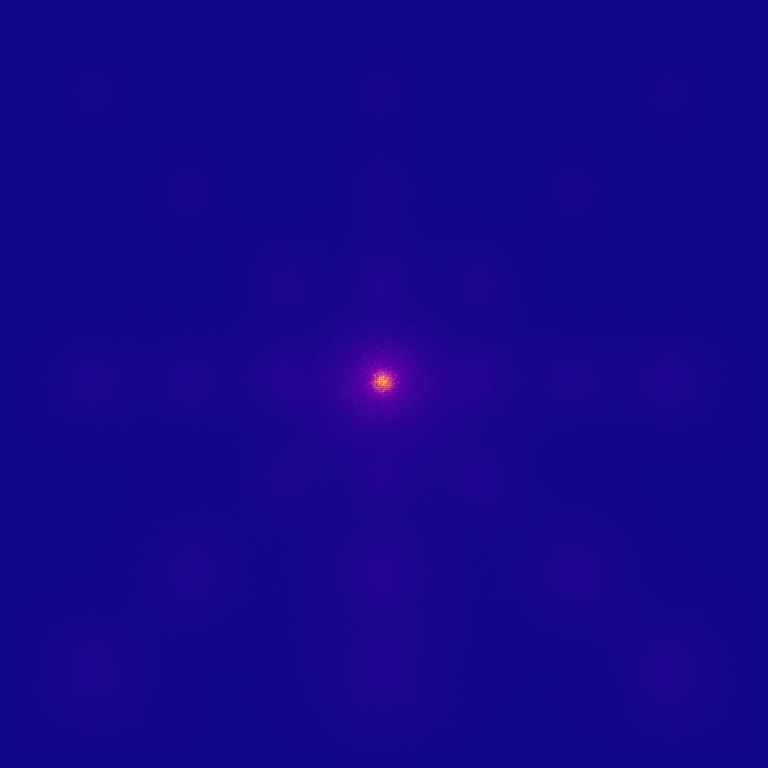"}
		\caption{DeepLabV3+ with R-50, $s=8$}
	\end{subfigure}
	\hfill
	\begin{subfigure}[b]{0.241\linewidth}
		\centering
		\includegraphics[width=\linewidth]{"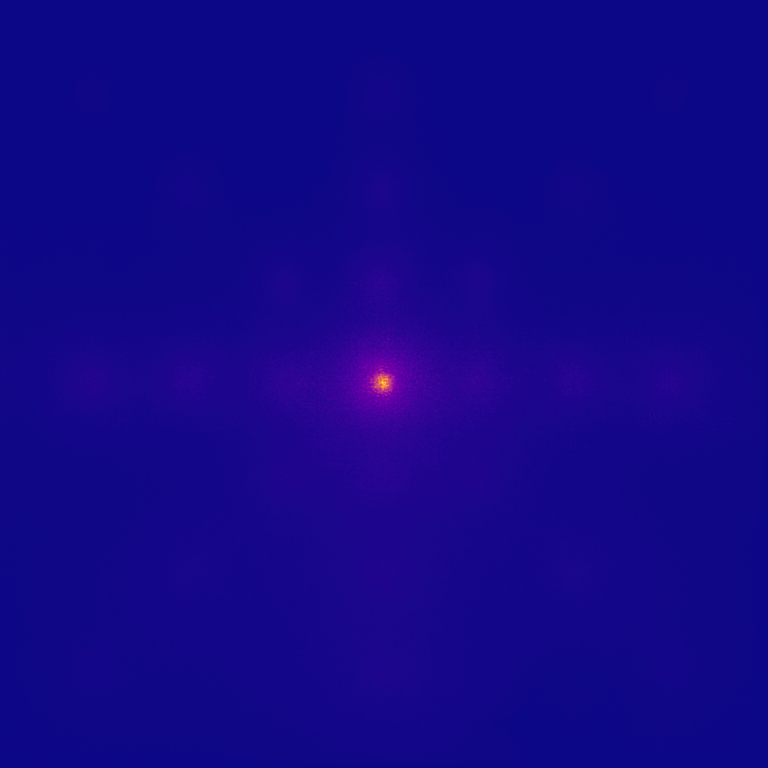"}
		\caption{DeepLabV3+ with R-101, $s=8$}
	\end{subfigure}
	\hfill
	\begin{subfigure}[b]{0.241\linewidth}
		\centering
		\includegraphics[width=\linewidth]{"./figure/configs_deeplabv3_deeplabv3_r101-d16-mg124_512x1024_80k_cityscapesplasma.png"}
		\caption{DeepLabV3+ with R-101, $s=16$}
	\end{subfigure}
	\caption{ERFs of DeepLabV3+ for the Cityscapes dataset in $768 \times 768$ input image.}
	\label{fig:d3p}
\end{figure}

\begin{figure}[t!]
	\centering
	\begin{subfigure}[b]{0.241\linewidth}
		\centering
		\includegraphics[width=\linewidth]{"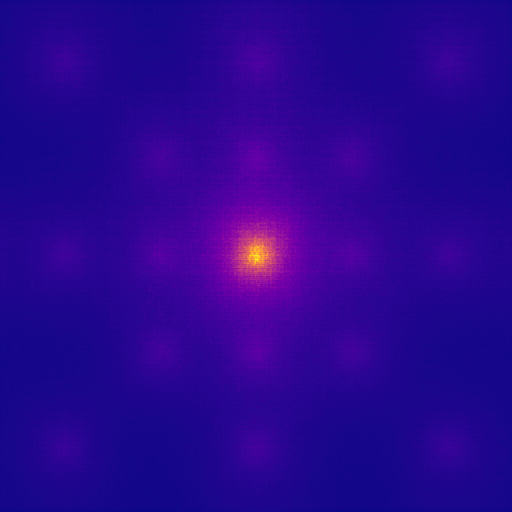"}
		\caption{DeepLabV3 with R-50, $s=8$}
	\end{subfigure}
	\hfill
	\begin{subfigure}[b]{0.241\linewidth}
		\centering
		\includegraphics[width=\linewidth]{"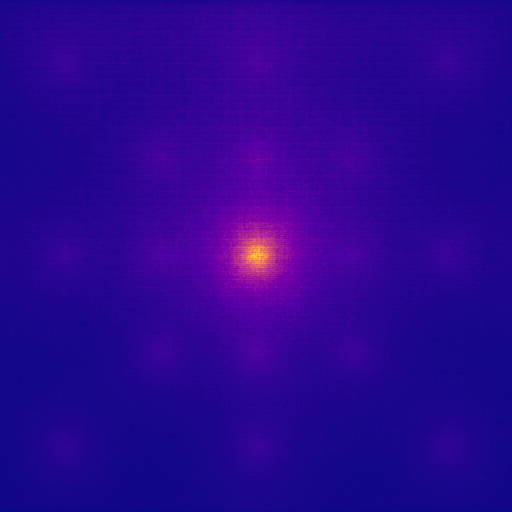"}
		\caption{DeepLabV3 with R-101, $s=8$}
	\end{subfigure}
	\hfill
	\begin{subfigure}[b]{0.241\linewidth}
		\centering
		\includegraphics[width=\linewidth]{"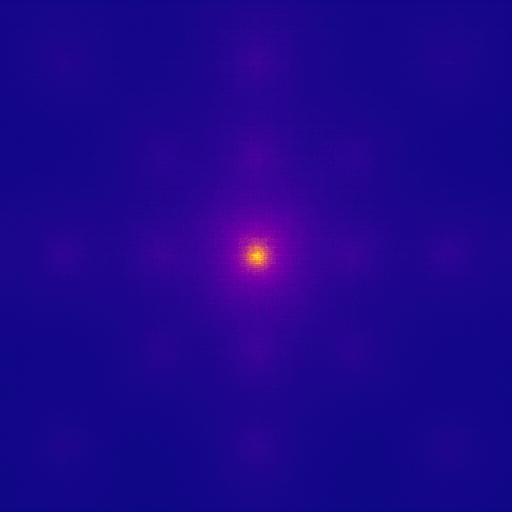"}
		\caption{DeepLabV3+ with R-50, $s=8$}
	\end{subfigure}
	\hfill
	\begin{subfigure}[b]{0.241\linewidth}
		\centering
		\includegraphics[width=\linewidth]{"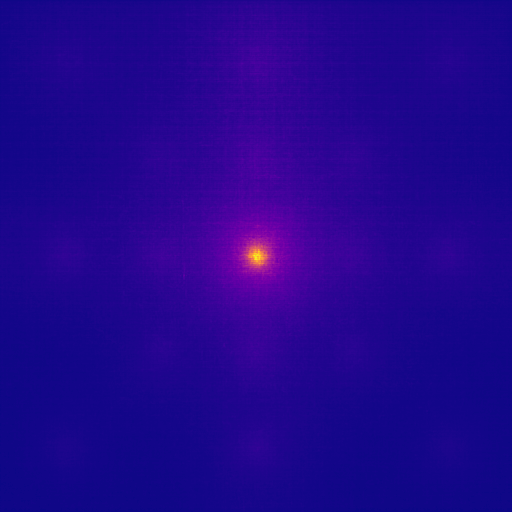"}
		\caption{DeepLabV3+ with R-101, $s=8$}
	\end{subfigure}
	\caption{ERFs of DeepLabV3 and DeepLabV3+ for the ADE20K dataset in $512 \times 512$ input image. Note that \EightStarTaper-shape appears like a cropped version of its larger size.}
	\label{fig:d3ade}
\end{figure}

\begin{figure}[t!]
	\centering
	\begin{subfigure}[b]{0.241\linewidth}
		\centering
		\includegraphics[width=\linewidth]{"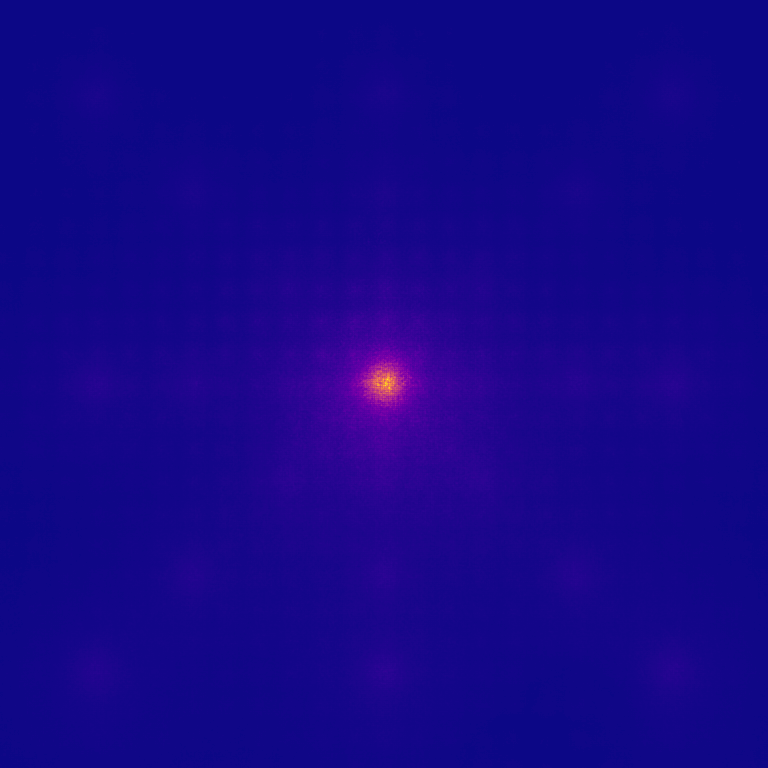"}
		\caption{FastFCN with ASPP head, $b=8$}
	\end{subfigure}
	\hfill
	\begin{subfigure}[b]{0.241\linewidth}
		\centering
		\includegraphics[width=\linewidth]{"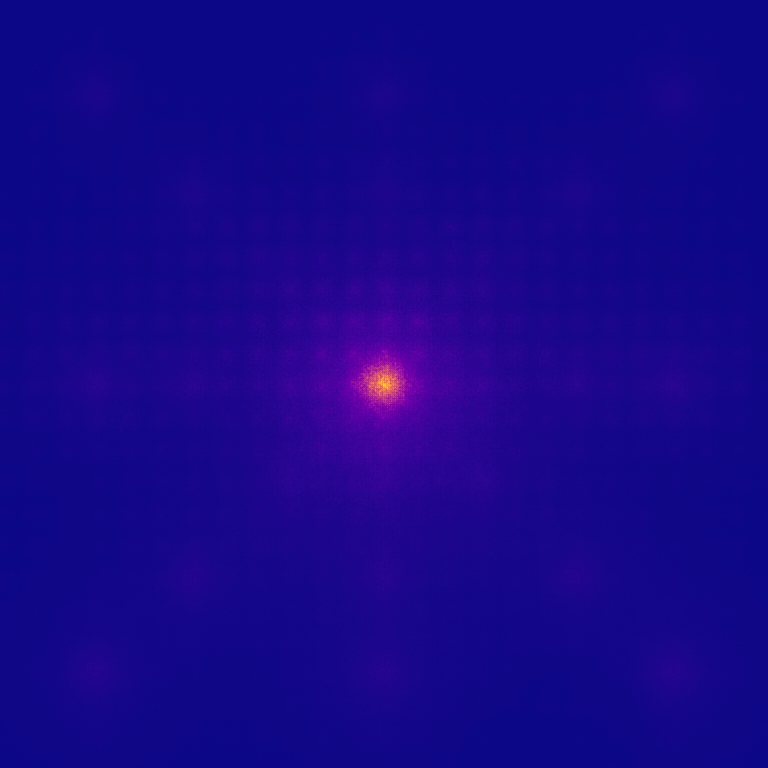"}
		\caption{FastFCN with ASPP head, $b=16$}
	\end{subfigure}
	\hfill
	\begin{subfigure}[b]{0.241\linewidth}
		\centering
		\includegraphics[width=\linewidth]{"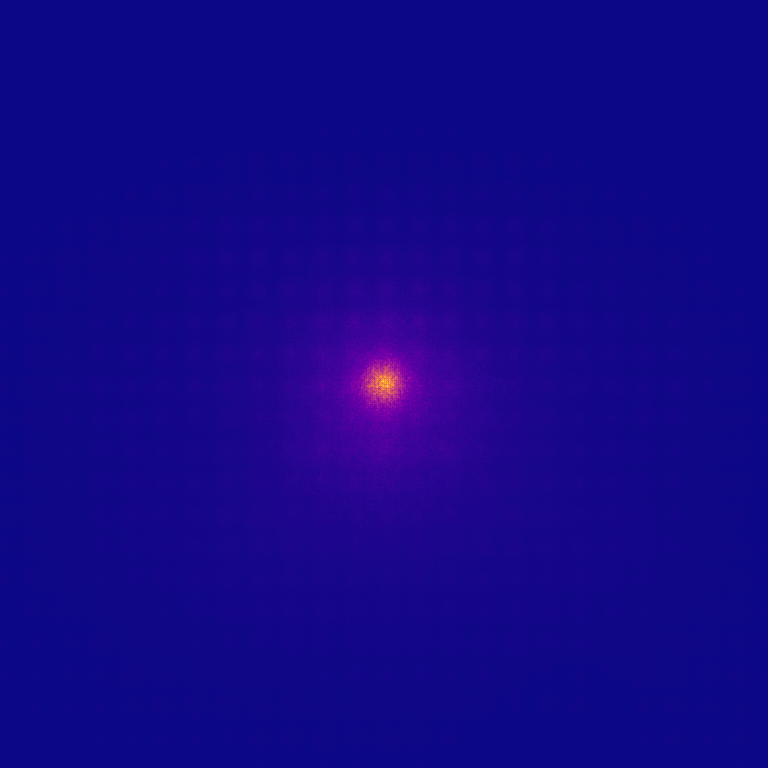"}
		\caption{FastFCN with PSP head, $b=8$}
	\end{subfigure}
	\hfill
	\begin{subfigure}[b]{0.241\linewidth}
		\centering
		\includegraphics[width=\linewidth]{"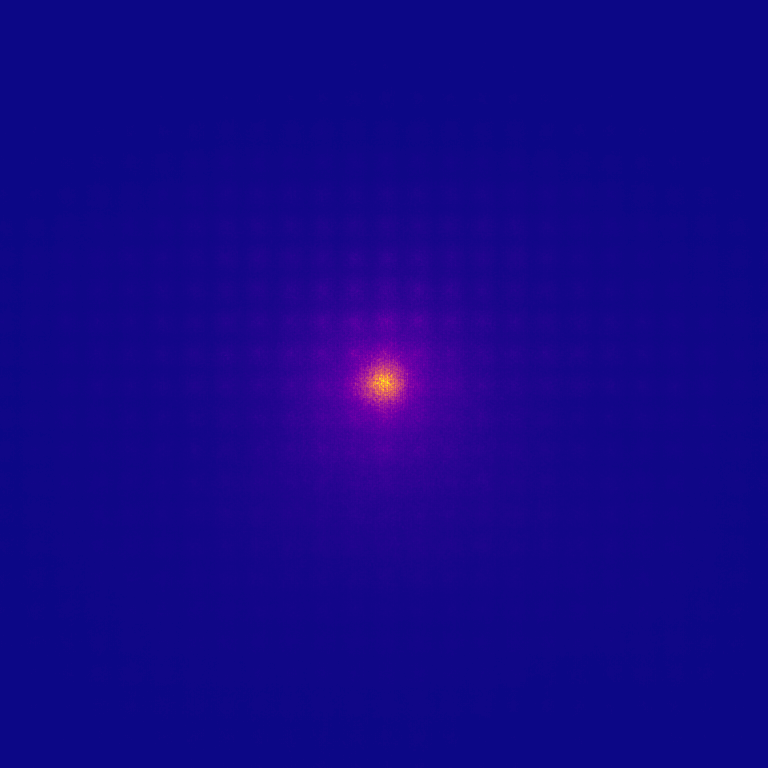"}
		\caption{FastFCN with PSP head, $b=16$}
	\end{subfigure}
	\caption{ERFs of FastFCN for the Cityscapes dataset in $768 \times 768$ input image. Notably, the \EightStarTaper-shape is visible in FastFCN with an ASPP head, but not in FastFCN with a PSP head. $b$ denotes the mini-batch size used during training.}
	\label{fig:fastfcn}
\end{figure}

Furthermore, the ERFs for DeepLabV3+ were obtained (\figref{fig:d3p}). Compared to DeepLabV3, DeepLabV3+ has a slightly different decoder that uses an additional branch to merge low-level features with the ASPP module. However, the ERFs of DeepLabV3+ still exhibited the star patterns. These observations imply that the star pattern is related to the use of the ASPP module.

Now, using another dataset called ADE20K \citep{DBLP:journals/ijcv/ZhouZPXFBT19}, the ERFs of DeepLabV3 and DeepLabV3+ were examined. Although cropped, the star pattern was visible in the ERFs (\figref{fig:d3ade}). The cropping occurred because of the $512 \times 512$ size of the input images. Compared to the Cityscapes dataset, the ADE20K dataset contains images that are smaller in size.

Finally, using another segmentation network called FastFCN \citep{DBLP:journals/corr/abs-1903-11816}, the ERFs for Cityscapes dataset were obtained. For FastFCN, the type of head, such as ASPP or PSP \citep{DBLP:conf/cvpr/ZhaoSQWJ17}, can be chosen. \figref{fig:fastfcn} illustrates the ERFs of FastFCNs using ASPP and PSP heads. FastFCN with an ASPP head exhibited a star pattern, whereas that with a PSP head did not.

Here, the observations from ERFs in Figures~\ref{fig:d3}--\ref{fig:fastfcn} are summarized.
\begin{itemize}
	\item \textbf{Observation 1.} When a segmentation network employed the ASPP module, its ERF exhibited the star pattern.
	\item \textbf{Observation 2.} When we follow the existing rule of atrous rate in \eqref{eq:bar}, the size of the star pattern did not expand or shrink by selecting the output stride as $s=8$ or $s=16$.
	\item \textbf{Observation 3.} The absolute size of the star pattern was fixed for current segmentation networks. When small-sized images were used, the star pattern in ERF was not shrunk but rather cropped.
\end{itemize}

Now, we analyze the underlying mechanism of the ASPP module to understand and validate these three observations.

\subsection{The Mechanism of ASPP Module}
\label{sec:howaspp}

\begin{figure}[t!]
	\centering
	\includegraphics[width=0.59\linewidth]{"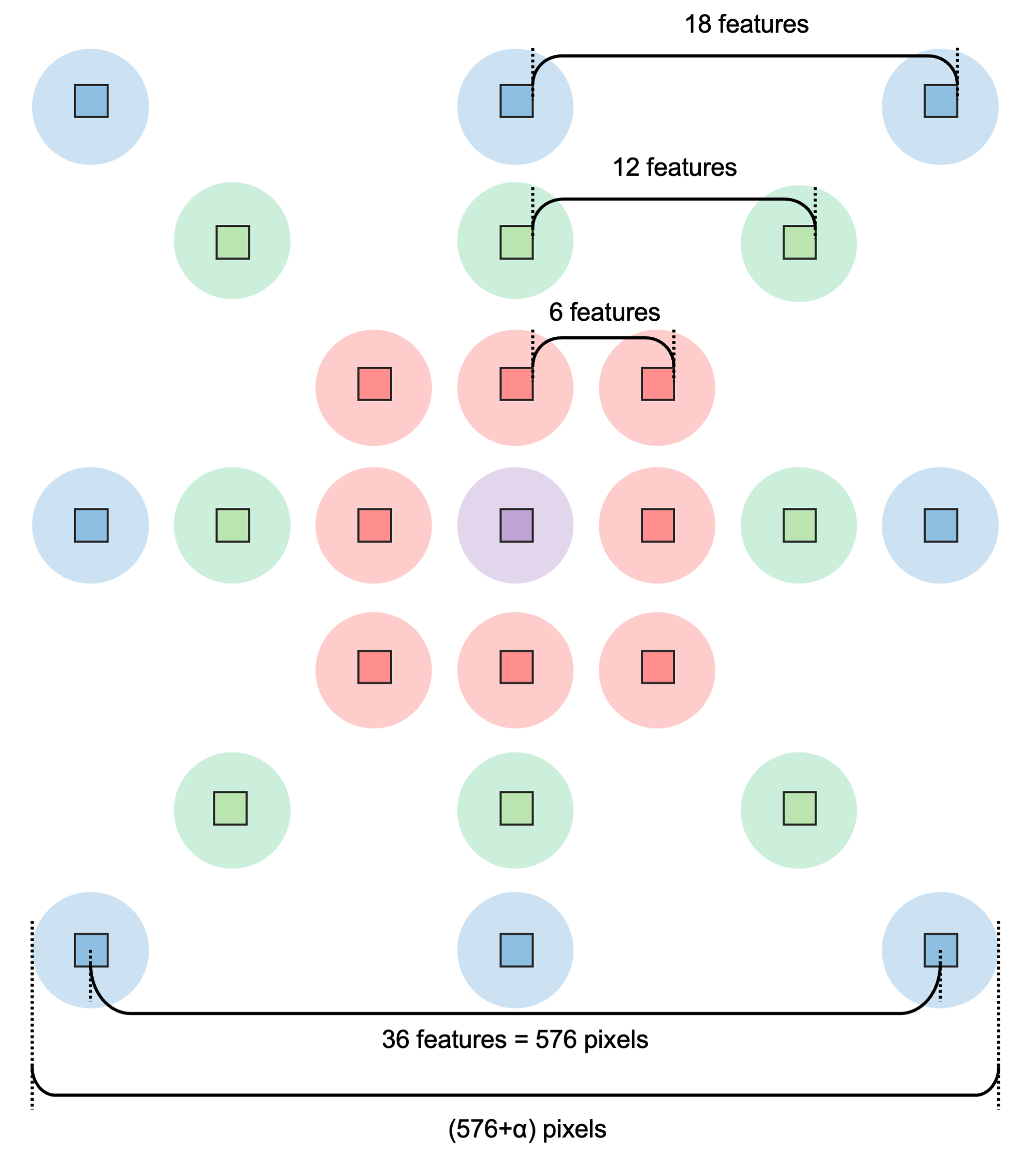"}
	\caption{Explanation of the FOV of ASPP module using atrous rates $\{6, 12, 18\}$ for an output stride $s=16$.}
	\label{fig:aspp}
\end{figure}

The ASPP module produces output $\bA$ using the encoded feature $\bH$. Here, we analyze the formation process of $\bA$ from $\bH$. Consider the base atrous rate $r=6$ and output stride $s=16$. The three atrous branches apply atrous convolutions parallelly to the feature map $\bH$ at rates $\{6, 12, 18\}$. The first $3 \times 3$ atrous convolution with an atrous rate of six uses nine features in $\bH$, and the distance between each feature is six in the feature unit (red boxes in \figref{fig:aspp}). Secondly, the $3 \times 3$ atrous convolution with an atrous rate of 12 uses one feature at the same target coordinate and eight features in $\bH$ at a distance of 12 on the feature unit (green boxes). Similarly, the $3 \times 3$ atrous convolution with an atrous rate of 18 uses one feature at the same target coordinate and eight features in $\bH$ with 18 feature distances (blue boxes). Here, each feature in the encoded feature $\bH$ covers a local area in the image $\bI$.

By the composition of the three atrous convolutions, each element of $\bA$ uses a single feature on the same target coordinate (purple), eight features at atrous rates 6 (red), 12 (green), and 18 (blue), amounting to $1+8+8+8=25$ features in $\bH$. Note that the 25 features in $\bH$ are regularly spaced, forming an \EightStarTaper-shape. This is the reason why employing the ASPP module yielded a star pattern on ERF (Observation 1).

Additionally, $s=16$ indicates that the feature map $\bH$ contains $16 \times$ downsampled information from the input image. Thus, in terms of scale, one feature unit on $\bH$ is equivalent to $16 \times 16$ pixels at the image level. This scale conversion allows us to analyze 25 features in a pixel unit. For instance, the center-to-center distance between the bottom-left and bottom-right features is 36 feature units, which can be converted to $36 \cdot 16 = 576$ pixels. When the center-to-center distance in ERF was measured using \figref{fig:d3}, approximately 573 pixels were obtained,\footnote{The distance was measured using a software such as XnViewMP.} that closely matched our expected value of 576 pixels.

Now the aforementioned analysis is generalized to $(r, s)$. The three atrous convolutions with atrous rates $\{r, 2r, 3r\}$ use 25 features on a regularly spaced star pattern. Furthermore, each feature unit in $\bH$ is equivalent to $s \times s$ pixel units at the image level. Thus, the center-to-center distance between the bottom-left and bottom-right features is $2 \cdot 3r$ feature units, which can be converted to $2 \cdot 3r \cdot s = 6rs$ pixels. When we consider the end-to-end distance between the bottom-left and bottom-right features, adding a margin $\alpha \approx 32$ of the circular areas, $6rs+\alpha$ pixels are obtained.

Note that when following the existing rule in \eqref{eq:bar}, two possible choices of $(r, s) = (6, 16)$ or $(r, s) = (12, 8)$ exist, resulting in the same length of $6rs=576$. This is the reason why both output strides $s=8$ and $s=16$ resulted in the same size of the star pattern (Observation 2). In other words, as long as we are limited to only two choices, the absolute size of star pattern is unaffected by other factors such as the dataset used (Observation 3).

\subsection{Guidelines for Determining the Atrous Rates of ASPP Module}
Based on the aforementioned analysis, we conclude that in order to enlarge the FOV of ASPP module, we should not adhere to the existing rule of atrous rates in \eqref{eq:bar} and should consider increasing either the base atrous rate $r$ or output stride $s$. However, a larger output stride implies a higher downsampling rate, which is disadvantageous for semantic segmentation. Furthermore, as discussed in the study by DeepLabV3 \citep{DBLP:journals/corr/ChenPSA17}, the choice of output stride can be limited by the available GPU resources. Therefore, rather than controlling the output stride, we would like to control the base atrous rate to enlarge or shrink the FOV of ASPP module.

\begin{figure}[t!]
	\centering
	\includegraphics[width=0.99\linewidth]{"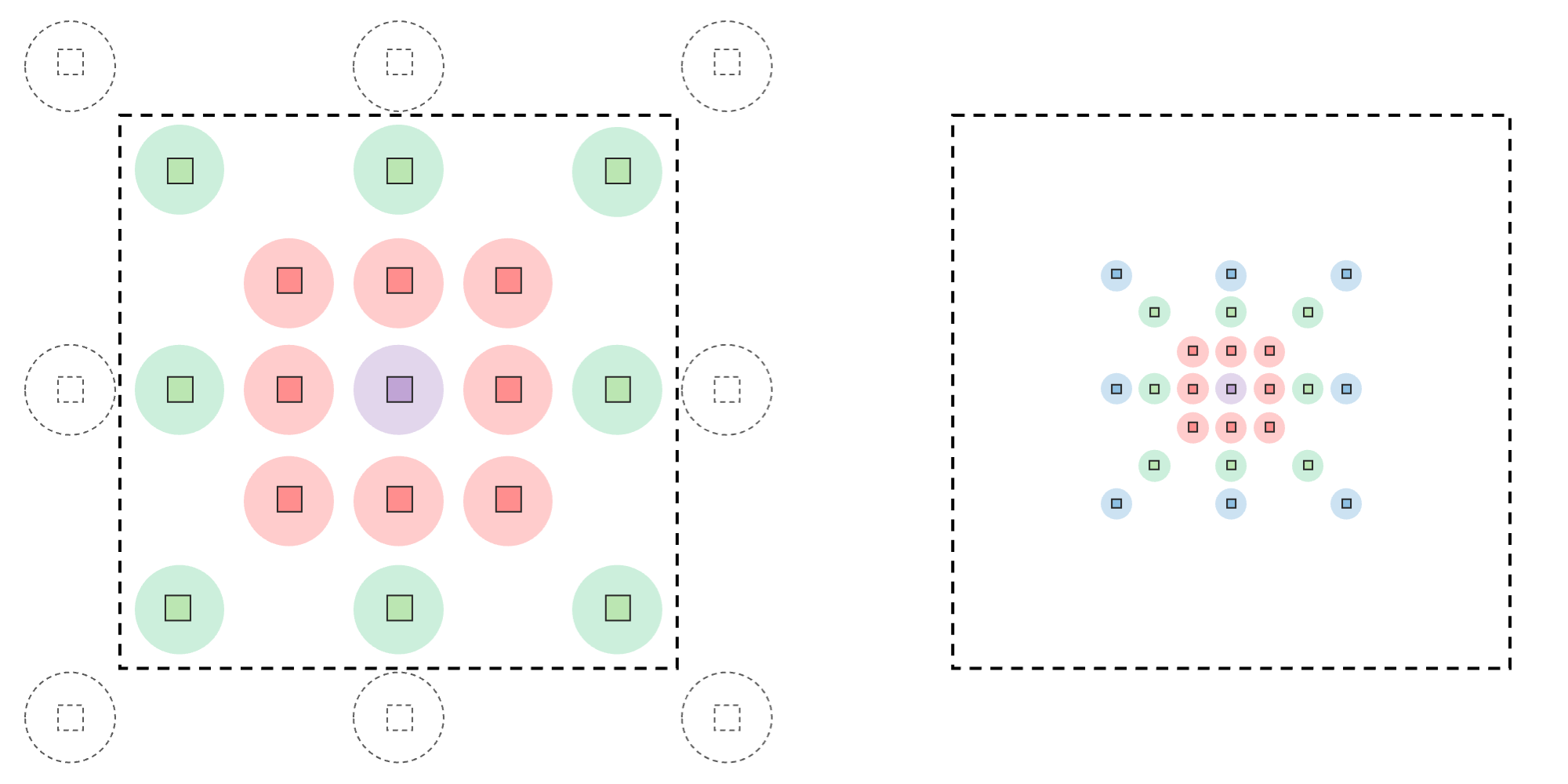"}
	\caption{Illustration comparing the FOV of ASPP (\EightStarTaper-shaped pattern) with image size (dotted line). (Left) When the FOV size of the ASPP module is larger than the image size, the outer kernel becomes an invalid operation. (Right) When the FOV size of the ASPP module is smaller than the image size, the segmentation network cannot capture global information in the image.}
	\label{fig:asppsize}
\end{figure}

Here, we claim that the FOV size of ASPP should match the size of the input image. Reference \citep{DBLP:journals/corr/ChenPSA17} demonstrated that if the FOV size of the ASPP module is larger than the size of input image, then the outer kernel in the $3 \times 3$ atrous convolution is applied to the zero-padded region, thereby causing an invalid kernel which degenerates into a $1 \times 1$ convolution. However, if the FOV size of ASPP module is smaller than the size of the input image, the ASPP module will not capture the global context of the image (\figref{fig:asppsize}). The limited usage of global information in an image is disadvantageous for semantic segmentation because performance of scene understanding is improved by using the global information \citep{DBLP:conf/cvpr/YangYZLY18}. Therefore, we regard the desirable size of FOV as the exact size of the input image, neither more than that nor less than that. Note that this size refers to that of the input of segmentation network, such as the crop size during training, not the size of the full image. In summary, our objective is to equalize the end-to-end distance $6rs+\alpha$ to the size of input image $l$. From this condition, we derive the following guidelines for obtaining the optimal base atrous rate $r^*$:
\begin{align}
	r^* = \frac{l-\alpha}{6s}. \label{eq:guide}
\end{align}

For instance, when performing semantic segmentation on the PASCAL VOC 2012 dataset, a crop size of $512 \times 512$ is commonly used. For $l=512$ and $s=16$, the proposed guidelines state that $r^*=5$, which is close to $r=6$ from the existing rule in \eqref{eq:bar}. We claim that this is the reason why the existing rule of atrous rates in \eqref{eq:bar} has worked suitably with the PASCAL VOC 2012 dataset. Furthermore, a crop size of $512 \times 512$ has been widely used for several datasets, including ADE20K, COCO Sfuff \citep{DBLP:conf/cvpr/CaesarUF18}, LoveDA \citep{DBLP:conf/nips/WangZMLZ21}, and REFUGE \citep{DBLP:journals/mia/OrlandoFBKBDFHK20}. In addition, $l=512$ was used in early studies on the Cityscapes dataset.

\begin{table}[t!]
	\caption{For the image size $l$ and output stride $s$, the proposed base atrous rate $r^*$ is summarized. In practice, the integer closest to $r^*$ is selected.}
	\label{tab:guide}
	\centering
	\begin{tabular}{rrr|rrr}
		\toprule
		size $l$ & stride $s$ & rate $r^*$ & size $l$ & stride $s$ & rate $r^*$ \\
		\midrule
		128      & 16         & 1.00       & 128      & 8          & 2.00       \\
		256      & 16         & 2.33       & 256      & 8          & 4.67       \\
		320      & 16         & 3.00       & 320      & 8          & 6.00       \\
		512      & 16         & 5.00       & 512      & 8          & 10.00      \\
		640      & 16         & 6.33       & 640      & 8          & 12.67      \\
		768      & 16         & 7.67       & 768      & 8          & 15.33      \\
		769      & 16         & 7.68       & 769      & 8          & 15.35      \\
		832      & 16         & 8.33       & 832      & 8          & 16.67      \\
		896      & 16         & 9.00       & 896      & 8          & 18.00      \\
		1024     & 16         & 10.33      & 1024     & 8          & 20.67      \\
		\bottomrule
	\end{tabular}
\end{table}

However, recent studies using the Cityscapes dataset have preferred a larger crop size, such as $769 \times 769$, which enables the aggregation of information on a wide image \citep{DBLP:conf/eccv/YinYCLZLH20,DBLP:journals/pami/HuangWWHSLH23,DBLP:journals/corr/abs-1907-12273,DBLP:conf/iccv/ZhuXBHB19}. For $l=769$ and $s=8$, the proposed guidelines state that $r^*=15.35$, which is different from $r=12$ in the existing rule in \eqref{eq:bar}. Indeed, the later section demonstrates that when using a crop size of $769 \times 769$ for the Cityscapes dataset, $r=15$ consistently yielded the highest performance in semantic segmentation among $r \in \{12, 13, \cdots, 18\}$. Conversely, studies on a few datasets have commonly used smaller crop sizes, such as $128 \times 128$, which require smaller FOV and atrous rates. In summary, we claim that the crop size of semantic segmentation can differ from $512 \times 512$; thus, a valid atrous rate should be used based on the proposed guidelines.

\section{Experiments}
\label{sec:exp}

\subsection{Small Image Size}

\begin{table}[t!]
	\caption{Results of semantic segmentation on the STARE dataset. The mIoU (\%) and its improvement $\Delta$ compared with the baseline ($r=12$) are reported. $^*$ indicates the base atrous rate according to the proposed guideline.}
	\label{tab:stare}
	\centering
    \begin{adjustbox}{width=\textwidth,center}
	\begin{tabular}{c|rrrrrrrrrrrr}
		\toprule
		$r$      & 12    & 11    & 10    & 9     & 8     & 7     & 6     & 5     & 4     & 3     & 2     & 1$^*$          \\
		\midrule
		mIoU     & 89.89 & 89.89 & 89.87 & 89.98 & 89.94 & 89.95 & 89.91 & 89.94 & 89.93 & 90.00 & 89.94 & \textbf{90.01} \\
		$\Delta$ & 0.00  & 0.00  & -0.02 & +0.09 & +0.05 & +0.06 & +0.02 & +0.05 & +0.04 & +0.11 & +0.05 & +0.12          \\
		\bottomrule
	\end{tabular}
    \end{adjustbox}
\end{table}

\begin{table}[t!]
	\caption{Results of semantic segmentation on the CHASE\_DB1 dataset. The mIoU (\%) and its improvement $\Delta$ compared with the baseline ($r=12$) are reported.}
	\label{tab:chase}
	\centering
    \begin{adjustbox}{width=\textwidth,center}
	\begin{tabular}{c|rrrrrrrrrrrr}
		\toprule
		$r$      & 12    & 11    & 10    & 9     & 8     & 7     & 6     & 5     & 4     & 3     & 2     & 1$^*$          \\
		\midrule
		mIoU     & 89.49 & 89.53 & 89.52 & 89.48 & 89.50 & 89.48 & 89.48 & 89.46 & 89.44 & 89.51 & 89.57 & \textbf{89.59} \\
		$\Delta$ & 0.00  & +0.04 & +0.03 & -0.01 & +0.01 & -0.01 & -0.01 & -0.03 & -0.05 & +0.02 & +0.08 & +0.10          \\
		\bottomrule
	\end{tabular}
    \end{adjustbox}
\end{table}

\begin{table}[t!]
	\caption{Results of semantic segmentation on the HRF dataset. The mIoU (\%) and its improvement $\Delta$ compared with the baseline ($r=12$) are reported.}
	\label{tab:hrf}
	\centering
    \begin{adjustbox}{width=\textwidth,center}
	\begin{tabular}{c|rrrrrrrrrrrr}
		\toprule
		$r$      & 12    & 11    & 10    & 9     & 8     & 7     & 6     & 5     & 4     & 3     & 2$^*$          & 1     \\
		\midrule
		mIoU     & 89.48 & 89.58 & 89.58 & 89.46 & 89.52 & 89.54 & 89.58 & 89.57 & 89.53 & 89.63 & \textbf{89.66} & 89.52 \\
		$\Delta$ & 0.00  & +0.10 & +0.10 & -0.02 & +0.04 & +0.06 & +0.10 & +0.09 & +0.05 & +0.15 & +0.18          & +0.04 \\
		\bottomrule
	\end{tabular}
    \end{adjustbox}
\end{table}

From this section, we compare the performance of segmentation networks with the proposed atrous rate and different values across various datasets and crop sizes. First, the performance of segmentation network using the ASPP module was examined with small crop sizes. We target the structured analysis of the retina (STARE) dataset \citep{DBLP:journals/tmi/HooverKG00,DBLP:journals/tmi/HooverG03}, which contains retinal images along with the corresponding segmentation labels on the blood vessels. Following the common practice for semantic segmentation of the STARE dataset \citep{DBLP:journals/tbe/YanYC18}, a crop size of $128 \times 128$ pixels was used, which was obtained after applying mean-std normalization and a random resize operation using a size of $605 \times 700$ pixels with a ratio range of 0.5 to 2.0. Furthermore, random flipping with a probability of 0.5 and photometric distortions, including brightness, contrast, saturation, and hue, were applied. The objective was to classify each pixel into one of the two categories and train the segmentation network using the cross-entropy loss function.

Following the common practice for semantic segmentation of the STARE dataset, U-Net \citep{DBLP:conf/miccai/RonnebergerFB15} using the ASPP module was targeted. U-Net employs four $2\times2$ maxpool operations, producing an output stride $s=16$. However, existing practices for U-Net with the ASPP module use the base atrous rate $r=12$ instead of $r=6$ from \eqref{eq:bar} proposed in DeepLabV3 and DeepLabV3+ \citep{mmseg2020}. To investigate the validity of the proposed atrous rate $r^*$ and compare it with the existing values, 12 different values of $r \in \{12, 11, \cdots, 1\}$ were experimented.

A training recipe from \texttt{MMSegmentation} \citep{mmseg2020} was employed. For training, stochastic gradient descent with momentum 0.9, weight decay $5 \times 10^{-4}$, and learning rate $10^{-2}$ with polynomial decay with a 40K scheduler were used. The training was performed on a single GPU machine. We measured the mean intersection over union (mIoU) and reported the average of ten runs.

\tabref{tab:stare} summarizes the results on the STARE dataset. Note that $r \in \{11, 10, \cdots, 4\}$ resulted in a similar mIoU to that of the baseline of $r=12$. In contrast, using $r \leq 3$ significantly increased the mIoU. These observations are consistent with the analysis of this study. When using a small size of input image, the invalid kernel activates as $r$ approaches a small value, thereby causing the ASPP module to operate effectively. The best mIoU was observed when $r=1$, which commensurates with the proposed guidelines stating that $r^*=1$ for $l=128$ and $s=16$.

To further validate the performance differences for small crop sizes, another dataset called the child heart and health study in england database (CHASE\_DB1) \citep{DBLP:journals/tbe/FrazRHUROB12} was targeted. Similar to the STARE dataset, the CHASE\_DB1 dataset contains retinal vessel images of children. Following the common practice for semantic segmentation of the CHASE\_DB1 dataset \citep{DBLP:journals/tbe/YanYC18}, a crop size of $128 \times 128$ pixels was used, which was obtained after applying mean-std normalization and a random resize operation with a size of $960 \times 999$ pixels in a ratio range of 0.5 to 2.0. In these experiments, the U-Net and training recipe similar to those for the STARE dataset were used.

\tabref{tab:chase} summarizes the results obtained from the average of three runs. Similarly, significant improvements were observed using $r \leq 2$. The best mIoU was observed at $r=1$, corresponding to the proposed base atrous rate $r^*=1$.

Finally, the high-resolution fundus (HRF) dataset \citep{DBLP:journals/ijbi/BudaiBMHM13} was targeted, which contains retinal fundus images with corresponding segmentation labels. Following the common practice for semantic segmentation of the HRF dataset \citep{DBLP:conf/midl/BreaJ0W20}, a crop size of $256 \times 256$ pixels was used, which was obtained after applying mean-std normalization and a random resize operation using a size of $2336 \times 3504$ pixels with a ratio range of 0.5 to 2.0. Furthermore, an additional Dice loss with a coefficient of 3.0 was used. In these experiments, the U-Net and training recipe similar to those for the STARE dataset were used.

\tabref{tab:hrf} summarizes the results obtained from the average of three runs. The best mIoU was observed at $r=2$, corresponding to the proposed guidelines stating that $r^*=2.33$ for $l=256$ and $s=16$.

\subsection{Large Image Size}
\label{sec:largesize}

\begin{table}[t!]
	\caption{Results of semantic segmentation on the Cityscapes dataset. The mIoU (\%) and its improvement $\Delta$ compared with the baseline ($r=12$) are reported.}
	\label{tab:city}
	\centering
	\begin{tabular}{l|c|rrrrrrr}
		\toprule
		Model                                 & $r$      & 12    & 13    & 14    & 15$^*$         & 16    & 17    & 18    \\
		\midrule
		\multirow{2}{*}{DeepLabV3 with R-50}  & mIoU     & 79.33 & 79.73 & 79.48 & \textbf{79.93} & 79.73 & 79.62 & 79.69 \\
		                                      & $\Delta$ & 0.00  & +0.40 & +0.15 & +0.60          & +0.40 & +0.29 & +0.36 \\
		\midrule
		\multirow{2}{*}{DeepLabV3 with R-101} & mIoU     & 78.88 & 79.46 & 79.75 & \textbf{79.90} & 79.85 & 79.87 & 79.60 \\
		                                      & $\Delta$ & 0.00  & +0.58 & +0.87 & +1.02          & +0.97 & +0.99 & +0.72 \\
		\bottomrule
	\end{tabular}
\end{table}

\begin{table}
	\caption{Results of semantic segmentation on the iSAID dataset. The mIoU (\%) and its improvement $\Delta$ compared with the baseline ($r=12$) are reported.}
	\label{tab:isaid}
	\centering
	\begin{tabular}{c|rrrrrrrrr}
		\toprule
		$r$      & 12    & 13    & 14    & 15    & 16    & 17    & 18$^*$         & 19    & 20    \\
		\midrule
		mIoU     & 66.28 & 66.41 & 66.91 & 66.68 & 66.51 & 66.96 & \textbf{67.03} & 66.72 & 66.86 \\
		$\Delta$ & 0.00  & +0.13 & +0.63 & +0.40 & +0.23 & +0.68 & +0.75          & +0.44 & +0.58 \\
		\bottomrule
	\end{tabular}
\end{table}

Thereafter, the performance of the segmentation network with the ASPP module was experimented using larger crop sizes. The Cityscapes dataset \citep{DBLP:conf/cvpr/CordtsORREBFRS16} containing images of urban street scenes was targeted. A crop size of $769 \times 769$ pixels was used, which was obtained after applying mean-std normalization and a random resize operation using a size of $2049 \times 1025$ pixels with a ratio range of 0.5 to 2.0. Furthermore, random flipping with a probability of 0.5 and the photometric distortions were applied. The objective was to classify each pixel into one of the 19 categories and train the segmentation network using the cross-entropy loss function.

We target DeepLabV3 that employs the ASPP module with an output stride $s=8$ and a default value of the base atrous rate $r=12$ from \eqref{eq:bar}. To investigate the validity of the proposed atrous rate $r^*$ and compare it with the existing value, seven different values of $r \in \{12, 13, \cdots, 18\}$ were experimentally investigated. Additionally, two backbones of ResNet-\{50, 101\} pretrained on ImageNet \citep{DBLP:conf/cvpr/DengDSLL009} were examined.

For training, stochastic gradient descent with momentum 0.9, weight decay $5 \times 10^{-4}$, and learning rate $10^{-2}$ with polynomial decay with an 80K scheduler were used. The training was conducted on a $4 \times$ GPU machine, and SyncBN \citep{DBLP:conf/cvpr/0005DSZWTA18} was used for distributed training. We measured the mIoU and reported the average of three runs.

\tabref{tab:city} summarizes the results on the Cityscapes dataset. Overall, $r \geq 13$ yielded improved mIoUs compared to the baseline performance of $r=12$. A peak improvement was observed using $r=15$, corresponding to the proposed guidelines stating that $r^*=15.35$ for $l=769$ and $s=8$. Although $r \geq 16$ yielded an improved mIoU, the degree of improvement decreased as the base atrous rate increased. This phenomenon was observed for the two backbones of ResNet-\{50, 101\}. These observations are consistent with our analysis. The FOV size of ASPP module should be controlled to exactly match the input size, neither more than that nor less than that.

To further validate the performance difference using large crop sizes, instance segmentation in aerial images dataset (iSAID) \citep{DBLP:conf/cvpr/ZamirAGKSK00XB19,DBLP:conf/cvpr/XiaBDZBLDPZ18} was targeted, which contains aerial images and the corresponding segmentation labels for object instances. Following the common practice for semantic segmentation of the iSAID dataset \citep{DBLP:conf/nips/GuoLHLC022}, an input size of $896 \times 896$ pixels was used, which was obtained after applying mean-std normalization and a random resize operation using a size of $896 \times 896$ pixels with a ratio range of 0.5 to 2.0. Furthermore, random flipping with a probability of 0.5 and the photometric distortions were used. The objective was to classify each pixel into one of the 16 categories and train the segmentation network using the cross-entropy loss function.

DeepLabV3+ that employs ASPP module and ResNet-50 with an output stride $s=8$ was targeted. In these experiments, the training recipe similar to that for the Cityscapes dataset was used. To investigate the validity of the proposed atrous rate $r^*$ and compare it with the existing value, nine different values of $r \in \{12, 13, \cdots, 20\}$ were experimented.

\tabref{tab:isaid} summarizes the results obtained from the average of two runs. Similarly, $r \geq 13$ yielded improved mIoUs compared to the baseline performance of $r=12$. The best mIoU was observed at $r=18$, corresponding to the proposed guidelines stating that $r^*=18$ for $l=896$ and $s=8$.

\begin{figure}[t!]
	\centering
	\begin{subfigure}[b]{0.191\linewidth}
		\centering
		\includegraphics[width=\linewidth]{"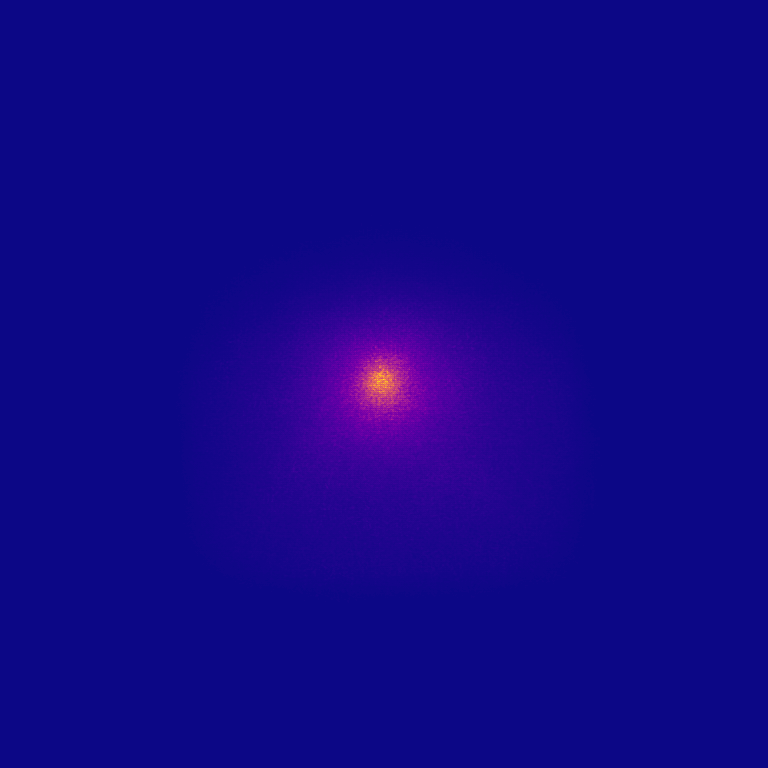"}
		\caption{FCN with R-50, $s=16$}
	\end{subfigure}
	\hfill
	\begin{subfigure}[b]{0.191\linewidth}
		\centering
		\includegraphics[width=\linewidth]{"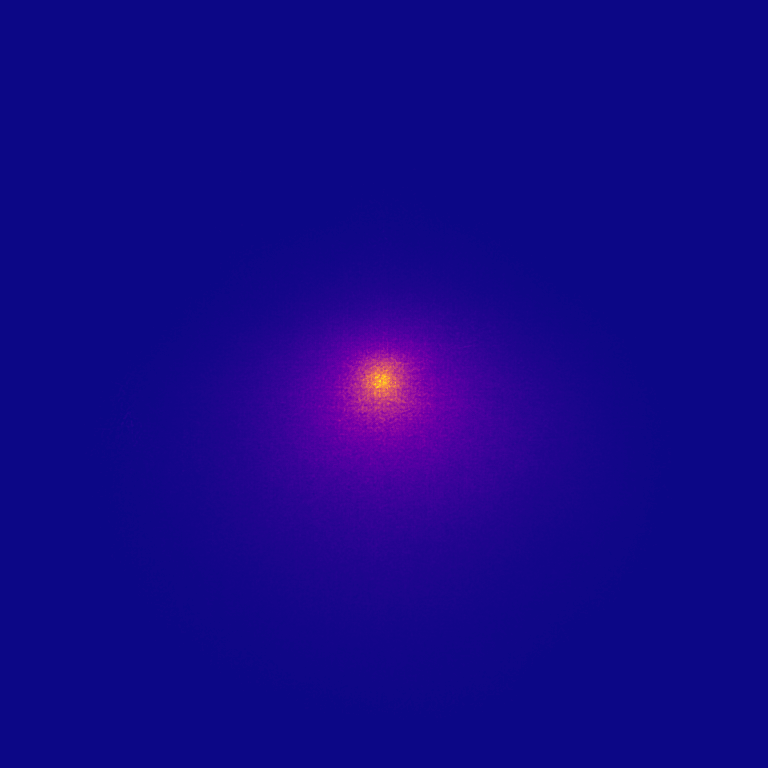"}
		\caption{FCN with R-101, $s=16$}
	\end{subfigure}
	\hfill
	\begin{subfigure}[b]{0.191\linewidth}
		\centering
		\includegraphics[width=\linewidth]{"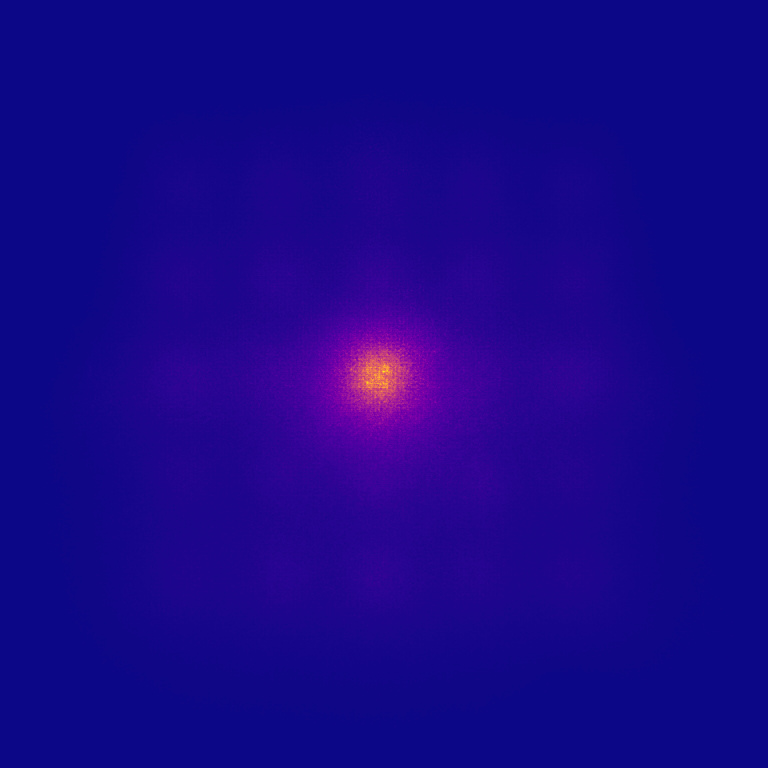"}
		\caption{FCN-D6 with R-50, $s=16$}
	\end{subfigure}
	\hfill
	\begin{subfigure}[b]{0.191\linewidth}
		\centering
		\includegraphics[width=\linewidth]{"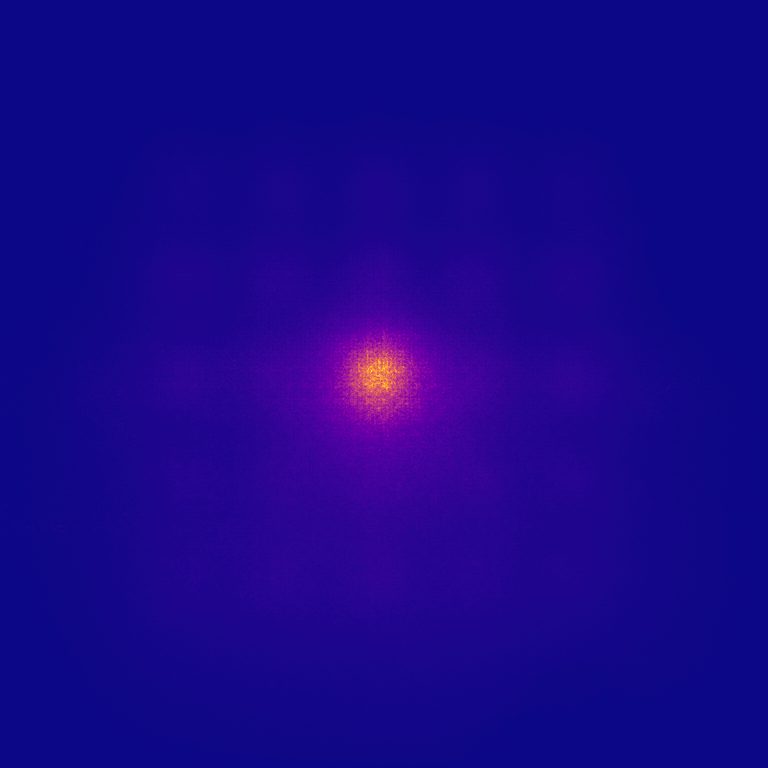"}
		\caption{FCN-D6 with R-101, $s=16$}
	\end{subfigure}
	\hfill
	\begin{subfigure}[b]{0.191\linewidth}
		\centering
		\includegraphics[width=\linewidth]{"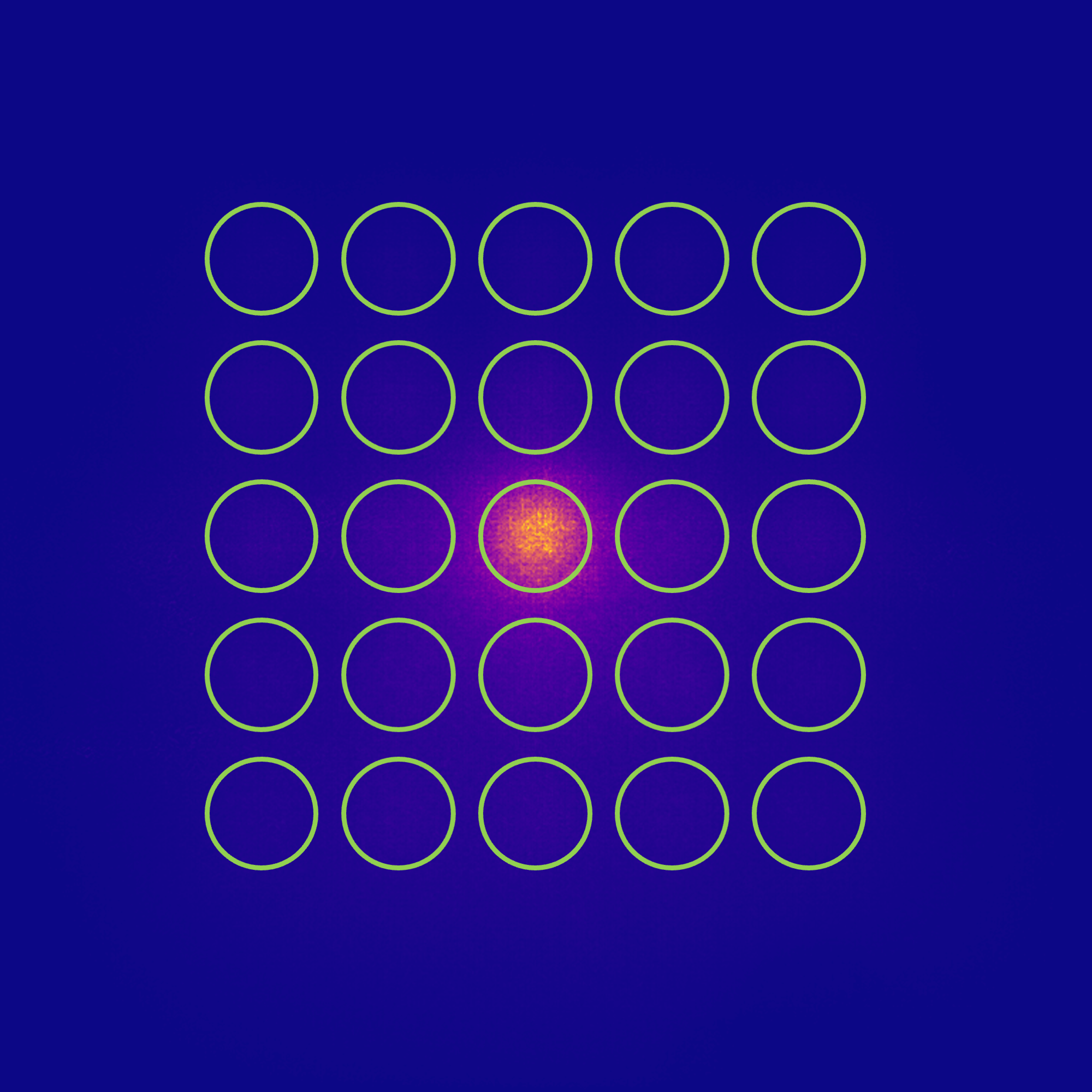"}
		\caption{Marked one for (d)}
	\end{subfigure}
	\caption{ERFs of FCN for the Cityscapes dataset in $768 \times 768$ input image. $5 \times 5$-shaped pattern is visible in FCN-D6 but not in the vanilla FCN.}
	\label{fig:fcn}
\end{figure}

\section{Discussion}
\label{sec:disc}

\subsection{Further Analysis of Various Segmentation Networks}
Here, ERFs for other segmentation networks are analyzed.

\paragraph{FCN and FCN-D6} FCN \citep{DBLP:journals/pami/ShelhamerLD17} is a representative segmentation network. Using ERF, FCN and its variant (\figref{fig:fcn}) were investigated. The vanilla FCN, which employs no atrous convolution in the head, yielded an ERF with a simple 2D Gaussian pattern. However, a variant called FCN-D6 yielded an ERF with a $5 \times 5$-shaped pattern, which was significantly different from a simple 2D Gaussian pattern. The ERF of FCN-D6 can be similarly explained using the analysis in \secref{sec:howaspp}. After obtaining the encoder output $\bH \in \R^{(H/s) \times (W/s) \times M}$ with $s=16$, FCN and FCN-D6 apply their heads to produce a semantic mask. Contrary to the vanilla FCN, the variant FCN-D6 employs a modified head with two $3 \times 3$ atrous convolutions with atrous rates of $\{6, 6\}$ in a row, not in parallel architecture. Note that applying the two $3 \times 3$ atrous convolutions with atrous rates $\{6, 6\}$ is equivalent to applying a single $5 \times 5$ atrous convolution with an atrous rate of six. Thus, the center-to-center distance between the bottom-left and bottom-right features is 24 feature units. Using an output stride of $s=16$, the center-to-center distance was converted into $24 \cdot 16 = 384$ pixels. Indeed, when measuring the center-to-center distance in \figref{fig:fcn}, approximately 379 pixels were obtained, which commensurates with the expected value. By generalizing into atrous rate $r$ and output stride $s$, the center-to-center distance of $4rs$ pixels is derived for FCN-D6. Therefore, when using FCN-D6, the FOV size should be controlled by setting a valid atrous rate $r$ depending on the target task or dataset.

\begin{figure}[t!]
	\centering
	\begin{subfigure}[b]{0.69\linewidth}
		\centering
		\includegraphics[width=\linewidth]{"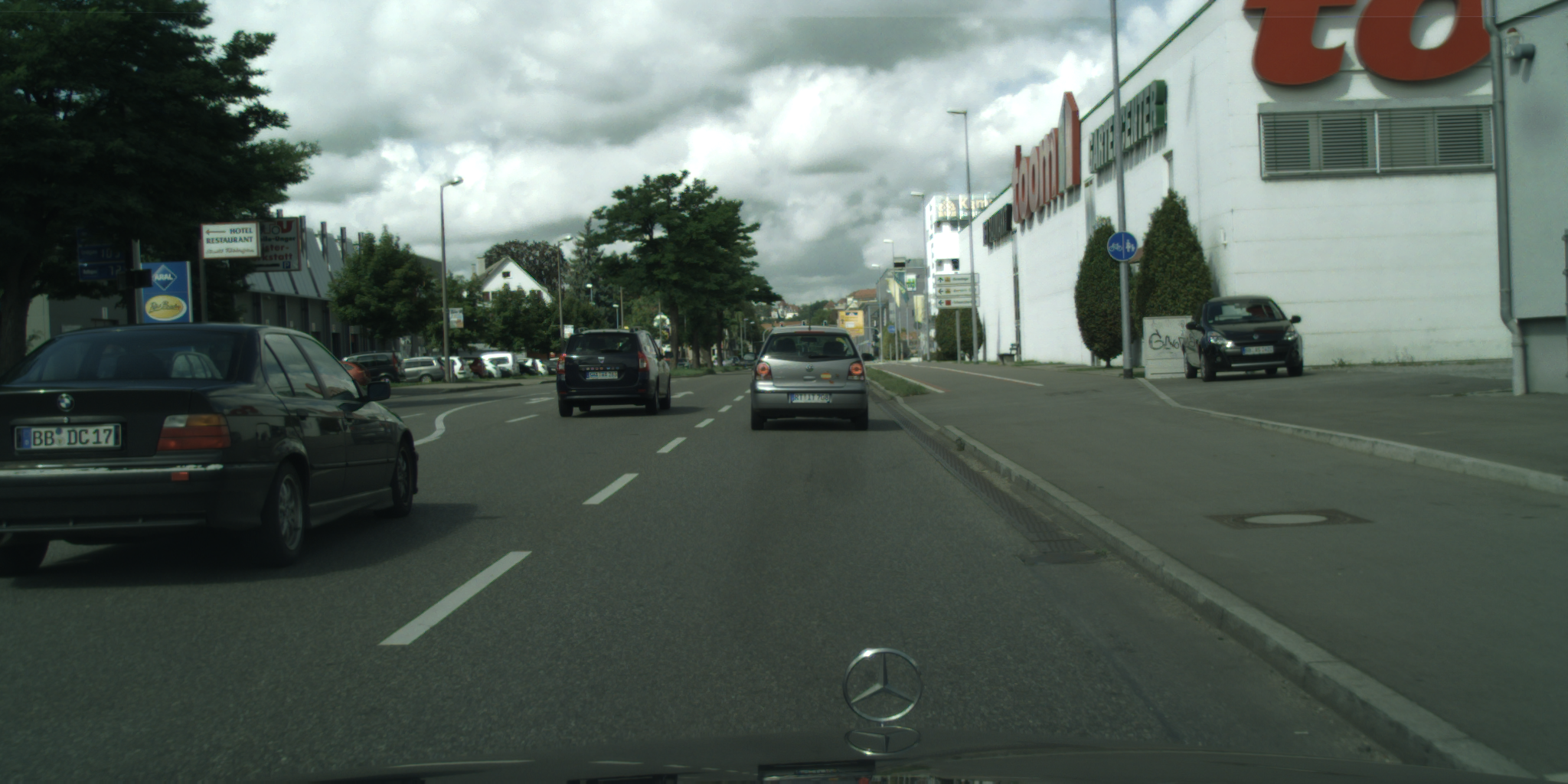"}
	\end{subfigure}
	\hfill
	\begin{subfigure}[b]{0.69\linewidth}
		\centering
		\includegraphics[width=\linewidth]{"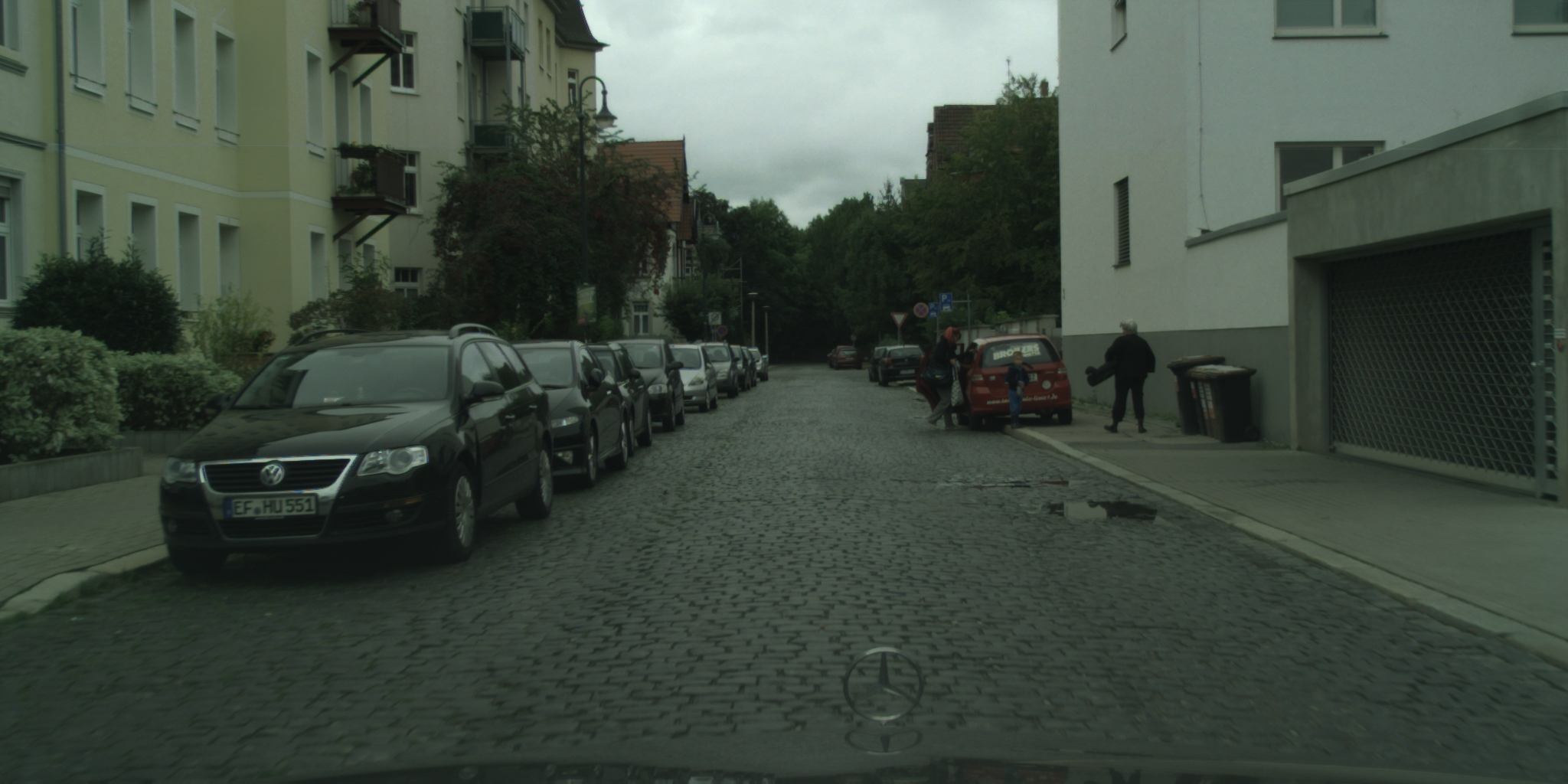"}
	\end{subfigure}
	\caption{Sample images of the Cityscapes dataset.}
	\label{fig:citysample}
\end{figure}

\begin{figure}[t!]
	\centering
	\begin{subfigure}[b]{0.241\linewidth}
		\centering
		\includegraphics[width=\linewidth]{"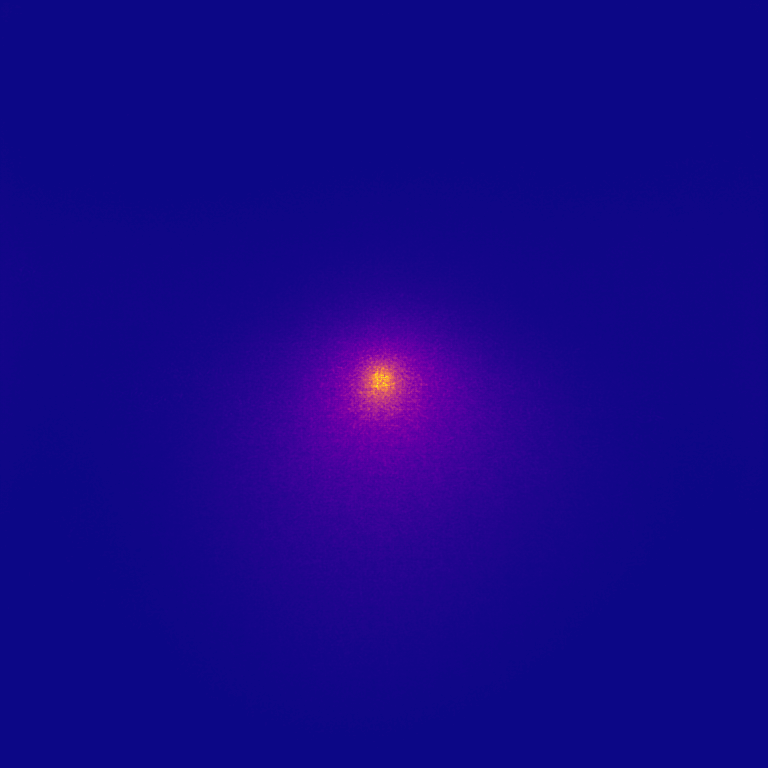"}
		\caption{EncNet \citep{DBLP:conf/cvpr/0005DSZWTA18}}
	\end{subfigure}
	\hfill
	\begin{subfigure}[b]{0.241\linewidth}
		\centering
		\includegraphics[width=\linewidth]{"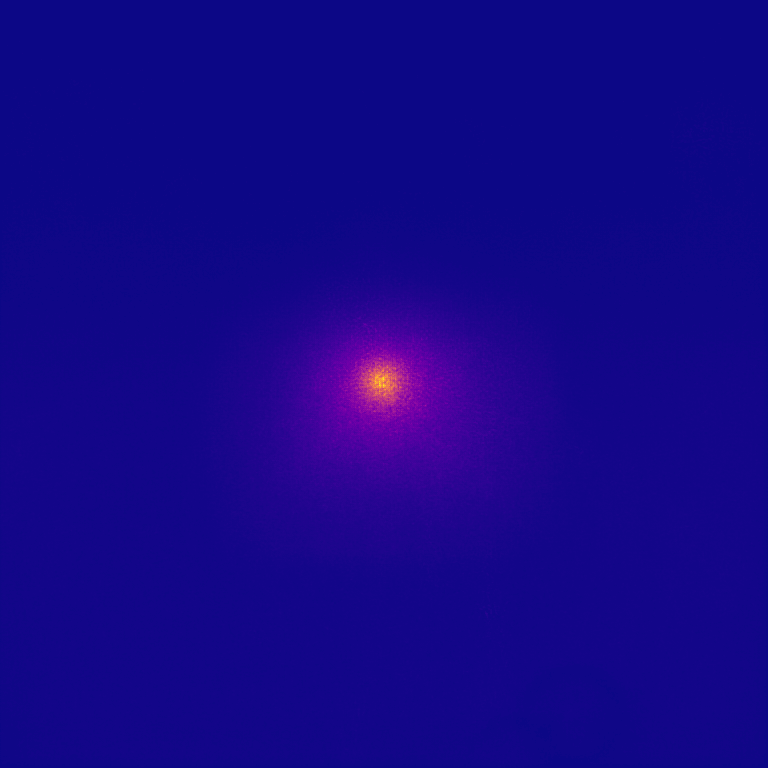"}
		\caption{NonLocalNet \citep{DBLP:conf/cvpr/0004GGH18}}
	\end{subfigure}
	\hfill
	\begin{subfigure}[b]{0.241\linewidth}
		\centering
		\includegraphics[width=\linewidth]{"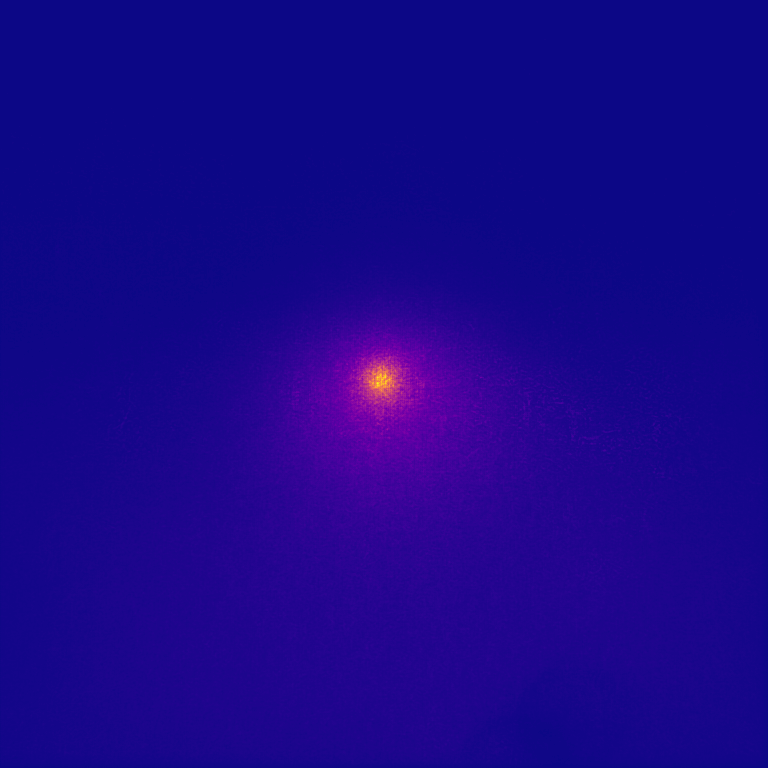"}
		\caption{ANN \citep{DBLP:conf/iccv/ZhuXBHB19}}
	\end{subfigure}
	\hfill
	\begin{subfigure}[b]{0.241\linewidth}
		\centering
		\includegraphics[width=\linewidth]{"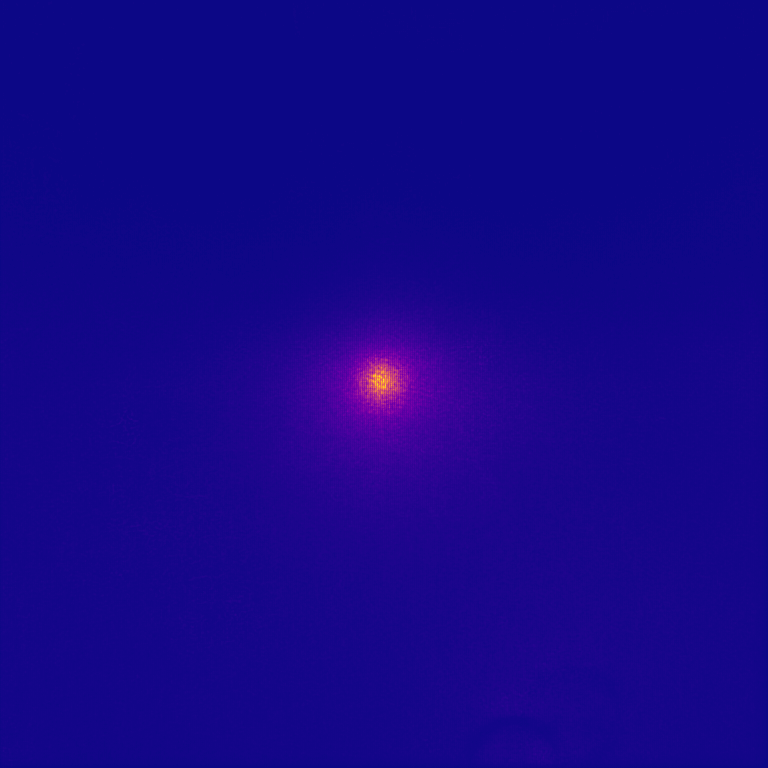"}
		\caption{DANet \citep{DBLP:conf/cvpr/FuLT0BFL19}}
	\end{subfigure}
	\caption{ERFs for the Cityscapes dataset in $768 \times 768$ input image. The use of global operations produces an ERF of an asymmetric 2D Gaussian pattern.}
	\label{fig:asy}
\end{figure}

\begin{table}[t!]
	\caption{Results of fitting each ERF to a 2D Gaussian for the Cityscapes dataset when the segmentation network employs global operations. The exact center coordinate is $(383, 384)$ for the $768 \times 768$ image. Although $x_c$ was closer to the center, $y_c$ was shifted to the bottom area. In addition, we observed $\sigma_x$ > $\sigma_y$, which indicates an asymmetric 2D Gaussian.}
	\label{tab:asy}
	\centering
	\begin{tabular}{l|rrrr}
		\toprule
		Model                                        & $x_c$ & $y_c$ & $\sigma_x$ & $\sigma_y$ \\
		\midrule
		EncNet \citep{DBLP:conf/cvpr/0005DSZWTA18}   & 382.8 & 397.4 & 58.1       & 55.0       \\
		NonLocalNet \citep{DBLP:conf/cvpr/0004GGH18} & 383.6 & 389.2 & 48.2       & 42.3       \\
		ANN \citep{DBLP:conf/iccv/ZhuXBHB19}         & 383.4 & 397.4 & 58.0       & 54.8       \\
		DANet \citep{DBLP:conf/cvpr/FuLT0BFL19}      & 382.3 & 385.7 & 40.4       & 35.1       \\
		\bottomrule
	\end{tabular}
\end{table}

\paragraph{Asymmetric Pattern for Cityscapes} The ERF of CNN is known as a symmetric 2D Gaussian \citep{DBLP:conf/nips/LuoLUZ16,DBLP:journals/prl/KimCJLJK23}. However, understanding a scene can require information on specific regions, depending on the dataset. In particular, for images in the Cityscapes dataset, distant objects in 3D space, such as the sky and buildings, are captured in the top area of the image, whereas closer objects and the structure of the road are positioned in the bottom area, whose information matters more in understanding the global context of the image (\figref{fig:citysample}). Consequently, the segmentation network for the Cityscapes dataset can prefer to focus on the bottom area of the image, yielding ERFs that exhibit an asymmetric 2D Gaussian pattern. We found that this phenomenon usually occurred when segmentation networks employed global modules, which used contextual information on all positions, such as the Context Encoding Module of EncNet \citep{DBLP:conf/cvpr/0005DSZWTA18}. In other words, the use of vanilla convolution restricts the segmentation network to having an ERF of symmetric 2D Gaussian, whereas employing global operations enables the segmentation network to flexibly choose an ERF pattern. \figref{fig:asy} shows the ERFs of asymmetric 2D Gaussian for the Cityscapes dataset, obtained from several segmentation networks that employ global modules. Indeed, NonLocalNet \citep{DBLP:conf/cvpr/0004GGH18} employs the non-local operation to aggregate features across all positions, and ANN \citep{DBLP:conf/iccv/ZhuXBHB19} is an improved version of NonLocalNet with a similar global operation. DANet \citep{DBLP:conf/cvpr/FuLT0BFL19} employs dual attention modules, which incorporate features with their global context. In addition, using \texttt{LMfit} library \citep{newville2016lmfit}, we fitted each ERF to a 2D Gaussian to obtain the center coordinates $(x_c, x_y)$ and the standard deviations $\sigma_x$ and $\sigma_y$, which represent the wideness of the 2D Gaussian (\tabref{tab:asy}). For four models, we observed that $x_c < y_c$ and $\sigma_x$ > $\sigma_y$, which indicates that these segmentation networks preferred and benefited from an asymmetric 2D Gaussian with a bottom-shifted and fat ERF to focus on road areas.

\begin{figure}[t!]
	\centering
	\begin{subfigure}[b]{0.241\linewidth}
		\centering
		\includegraphics[width=\linewidth]{"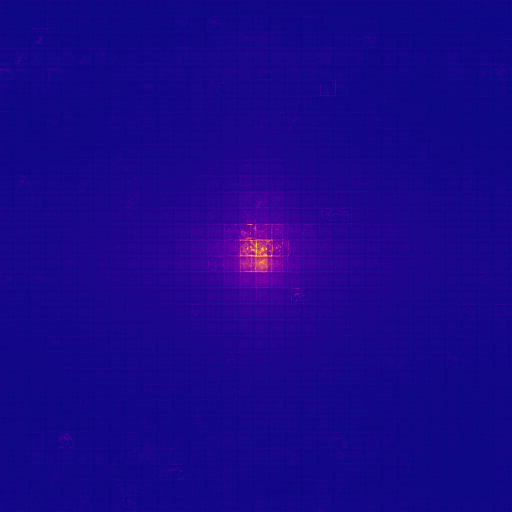"}
		\caption{ViT-B/16 \citep{DBLP:conf/iclr/DosovitskiyB0WZ21}}
	\end{subfigure}
	\hfill
	\begin{subfigure}[b]{0.241\linewidth}
		\centering
		\includegraphics[width=\linewidth]{"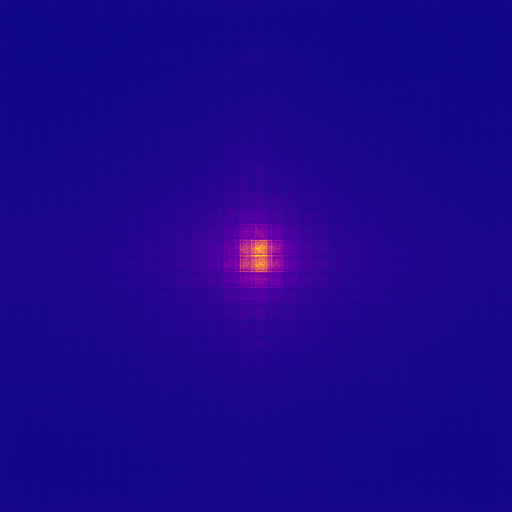"}
		\caption{DeiT-S/16 \citep{DBLP:conf/icml/TouvronCDMSJ21}}
	\end{subfigure}
	\hfill
	\begin{subfigure}[b]{0.241\linewidth}
		\centering
		\includegraphics[width=\linewidth]{"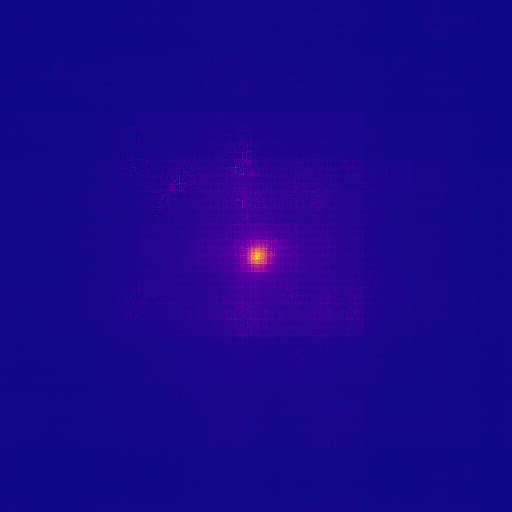"}
		\caption{Swin-S, W7 \citep{DBLP:conf/iccv/LiuL00W0LG21}}
	\end{subfigure}
	\hfill
	\begin{subfigure}[b]{0.241\linewidth}
		\centering
		\includegraphics[width=\linewidth]{"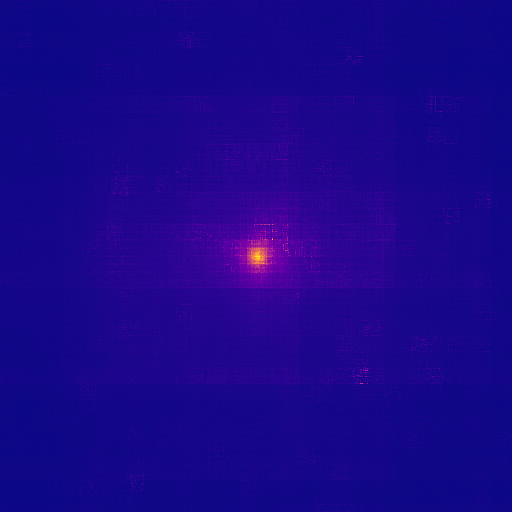"}
		\caption{Swin-B, W12 \citep{DBLP:conf/iccv/LiuL00W0LG21}}
	\end{subfigure}
	\caption{ERFs for the ADE20K dataset in $512 \times 512$ input image. UperNet with a transformer backbone. ``W'' indicates window size.}
	\label{fig:trback}
\end{figure}

\paragraph{Transformer Backbone} It is worthy to examine the properties of segmentation networks employing a recent vision transformer backbone. \figref{fig:trback} represents the ERFs of UperNet \citep{DBLP:conf/eccv/XiaoLZJS18} with transformer backbones. ViT-B/16 indicates the vision transformer at its base size and patch size of $16 \times 16$ pixels. Because ViT-B/16 \citep{DBLP:conf/iclr/DosovitskiyB0WZ21} partitions an image into a sequence of patches of $16 \times 16$ size, the ERF of UperNet with ViT-B/16 backbone exhibited patterns like patch-partitioned 2D Gaussian. UperNet with DeiT-S/16 \citep{DBLP:conf/icml/TouvronCDMSJ21} has an almost identical architecture to that of ViT with a smaller model size and yielded similar ERF. In contrast, the ERF of UperNet with Swin \citep{DBLP:conf/iccv/LiuL00W0LG21} exhibited a smooth 2D Gaussian without the patch-partitioned pattern because Swin merges features of neighboring patches.

\begin{figure}[t!]
	\centering
	\begin{subfigure}[b]{0.241\linewidth}
		\centering
		\includegraphics[width=\linewidth]{"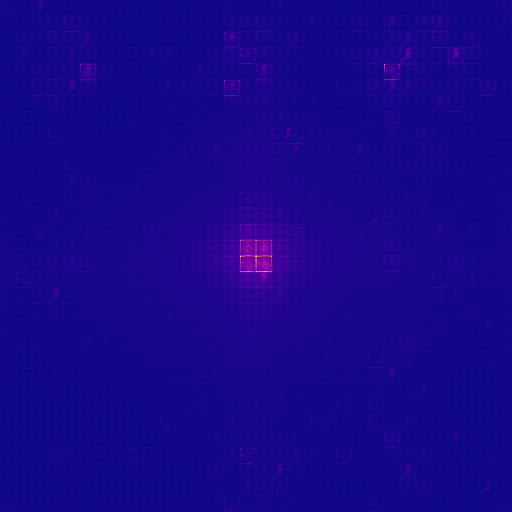"}
		\caption{SETR}
	\end{subfigure}
	\hfill
	\begin{subfigure}[b]{0.241\linewidth}
		\centering
		\includegraphics[width=\linewidth]{"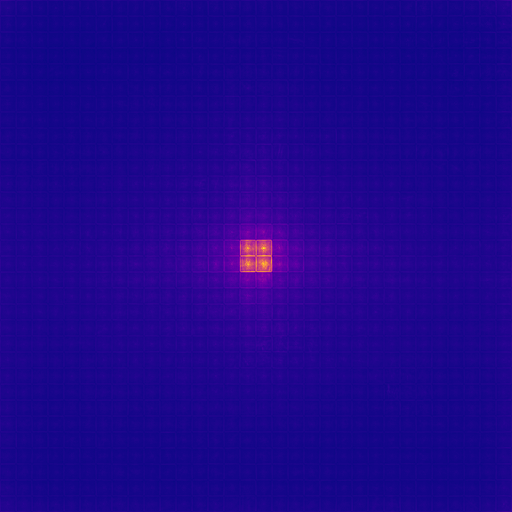"}
		\caption{Segmenter}
	\end{subfigure}
	\hfill
	\begin{subfigure}[b]{0.241\linewidth}
		\centering
		\includegraphics[width=\linewidth]{"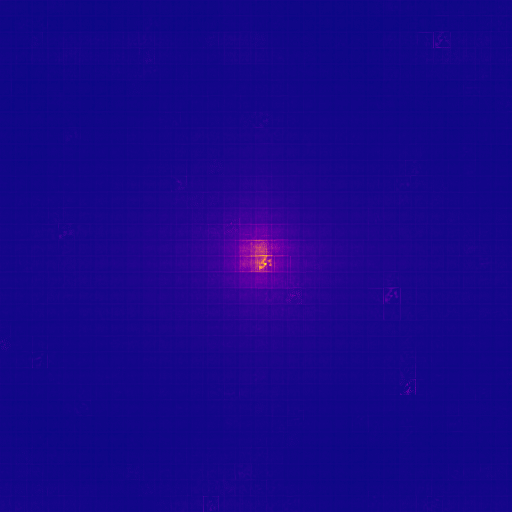"}
		\caption{DPT}
	\end{subfigure}
	\hfill
	\begin{subfigure}[b]{0.241\linewidth}
		\centering
		\includegraphics[width=\linewidth]{"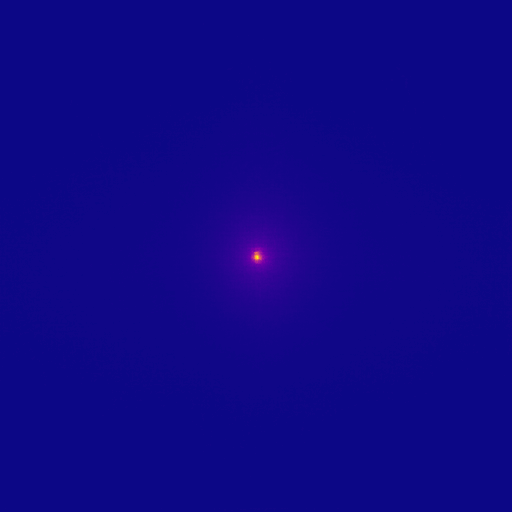"}
		\caption{SegFormer}
	\end{subfigure}
	\caption{ERFs for the ADE20K dataset in $512 \times 512$ input image.}
	\label{fig:trss}
\end{figure}

\paragraph{Transformer Models} In addition to using vision transformers as the backbone, several segmentation networks have been proposed to elaborately incorporate the self-attention mechanism of transformers. \figref{fig:trss} illustrates the ERFs of recent segmentation networks that employ the self-attention mechanism. For SETR \citep{DBLP:conf/cvpr/ZhengLZZLWFFXT021}, Segmenter \citep{DBLP:conf/iccv/StrudelPLS21}, and DPT \citep{DBLP:conf/iccv/RanftlBK21}, which employ ViT backbones, ERFs exhibited patch-partitioned 2D Gaussian patterns, but with different wideness due to their distinct decoder designs. Interestingly, SegFormer \citep{DBLP:conf/nips/XieWYAAL21}, which merges features of patches, exhibited ERF highlighting a significantly small area, while achieving high performance in semantic segmentation. This observation indicates that, although using global information from a wide ERF could be advantageous for semantic segmentation, certain segmentation networks rather focused on local information with smaller ERFs, while still achieving high performance.

\subsection{Transformer with ASPP Module} It is worthy to examine the combination of the ASPP module and segmentation network using the self-attention mechanism. Here, SETR naive that employs ViT-L/16 was targeted. Note that the patch size 16 indicates output stride $s=16$. A crop size of $768 \times 768$ pixels was used. In these experiments, the same training recipe similar to that for the Cityscapes dataset in \secref{sec:largesize} was used. To investigate the validity of the proposed atrous rate $r^*$ and compare it with other values, 12 different values of $r \in \{1, 2, \cdots, 12\}$ were experimented. In addition, the baseline performance of SETR was measured. \tabref{tab:setr} summarizes the results obtained from the average of three runs. The peak improvement was observed when $r=8$. Indeed, the proposed guidelines state that $r^*=8$ for $l=768$ and $s=16$. Nevertheless, using other values of $r$ can exhibit decreased mIoU, which means that when using the ASPP module for performance gain, a valid value of the base atrous rate should be used.

\begin{table}[t!]
	\caption{Results of semantic segmentation on the Cityscapes dataset using SETR. The mIoU (\%) and its improvement $\Delta$ compared with the baseline (SETR naive) are reported.}
	\label{tab:setr}
	\centering
    \begin{adjustbox}{width=\textwidth,center}
	\begin{tabular}{c|rrrrrrrrrrrrr}
		\toprule
		$r$      & Baseline & 1     & 2     & 3     & 4     & 5     & 6     & 7     & 8$^*$          & 9     & 10    & 11    & 12    \\
		\midrule
		mIoU     & 77.54    & 77.53 & 77.69 & 77.51 & 77.22 & 77.64 & 77.49 & 77.62 & \textbf{77.88} & 77.73 & 77.82 & 77.68 & 77.57 \\
		$\Delta$ & 0.00     & -0.01 & +0.15 & -0.03 & -0.32 & +0.10 & -0.05 & +0.08 & +0.34          & +0.19 & +0.28 & +0.14 & +0.03 \\
		\bottomrule
	\end{tabular}
    \end{adjustbox}
\end{table}

\section{Conclusion}
\label{sec:con}
In this study, the mechanism of the ASPP module were analyzed. We introduced and obtained ERFs of segmentation networks to empirically observe the inner behavior of the ASPP module. We found that the use of the ASPP module led to a star-shaped pattern on ERF. Based on these observations, we explained the mechanisms of the ASPP module and quantified its FOV size. To obtain effective behavior of the ASPP module, we suggested adjusting the FOV size to exactly match the size of the input image, proposing guidelines to obtain the optimal atrous rates. The validity of the proposed guidelines was examined across different datasets and image sizes. We observed that the proposed atrous rate consistently yielded improved mIoU compared with other values of the atrous rate.

\bibliographystyle{unsrt}
\bibliography{sample-base}

\begin{thebibliography}{10}

\bibitem{DBLP:journals/tog/AksoyOPPM18}
Yagiz Aksoy, Tae{-}Hyun Oh, Sylvain Paris, Marc Pollefeys, and Wojciech
  Matusik.
\newblock {Semantic soft segmentation}.
\newblock {\em {ACM} Trans. Graph.}, 37(4):72, 2018.

\bibitem{DBLP:journals/tog/HuangFL14}
Zhe Huang, Hongbo Fu, and Rynson W.~H. Lau.
\newblock {Data-driven segmentation and labeling of freehand sketches}.
\newblock {\em {ACM} Trans. Graph.}, 33(6):175:1--175:10, 2014.

\bibitem{DBLP:journals/tog/XiaoFZLQ09}
Jianxiong Xiao, Tian Fang, Peng Zhao, Maxime Lhuillier, and Long Quan.
\newblock {Image-based street-side city modeling}.
\newblock {\em {ACM} Trans. Graph.}, 28(5):114, 2009.

\bibitem{DBLP:journals/tog/ZhuAFW21}
Peihao Zhu, Rameen Abdal, John Femiani, and Peter Wonka.
\newblock {Barbershop: GAN-based image compositing using segmentation masks}.
\newblock {\em {ACM} Trans. Graph.}, 40(6):215:1--215:13, 2021.

\bibitem{DBLP:journals/tog/SchneiderT16}
Ros{\'{a}}lia~G. Schneider and Tinne Tuytelaars.
\newblock {Example-Based Sketch Segmentation and Labeling Using CRFs}.
\newblock {\em {ACM} Trans. Graph.}, 35(5):151:1--151:9, 2016.

\bibitem{DBLP:journals/pami/ChenPKMY18}
Liang{-}Chieh Chen, George Papandreou, Iasonas Kokkinos, Kevin Murphy, and
  Alan~L. Yuille.
\newblock {DeepLab: Semantic Image Segmentation with Deep Convolutional Nets,
  Atrous Convolution, and Fully Connected CRFs}.
\newblock {\em {IEEE} Trans. Pattern Anal. Mach. Intell.}, 40(4):834--848,
  2018.

\bibitem{DBLP:conf/cvpr/ChenPWXLSF0SOLL19}
Kai Chen, Jiangmiao Pang, Jiaqi Wang, Yu~Xiong, Xiaoxiao Li, Shuyang Sun,
  Wansen Feng, Ziwei Liu, Jianping Shi, Wanli Ouyang, Chen~Change Loy, and
  Dahua Lin.
\newblock {Hybrid Task Cascade for Instance Segmentation}.
\newblock In {\em {CVPR}}, pages 4974--4983, 2019.

\bibitem{DBLP:journals/corr/abs-2011-11675}
Liang{-}Chieh Chen, Huiyu Wang, and Siyuan Qiao.
\newblock {Scaling Wide Residual Networks for Panoptic Segmentation}.
\newblock {\em CoRR}, abs/2011.11675, 2020.

\bibitem{DBLP:conf/cvpr/ChengCZ0HAC20}
Bowen Cheng, Maxwell~D. Collins, Yukun Zhu, Ting Liu, Thomas~S. Huang, Hartwig
  Adam, and Liang{-}Chieh Chen.
\newblock {Panoptic-DeepLab: {A} Simple, Strong, and Fast Baseline for
  Bottom-Up Panoptic Segmentation}.
\newblock In {\em {CVPR}}, pages 12472--12482, 2020.

\bibitem{DBLP:journals/corr/abs-1907-10326}
Jin~Han Lee, Myung{-}Kyu Han, Dong~Wook Ko, and Il~Hong Suh.
\newblock {From Big to Small: Multi-Scale Local Planar Guidance for Monocular
  Depth Estimation}.
\newblock {\em CoRR}, abs/1907.10326, 2019.

\bibitem{DBLP:conf/cvpr/FuGWBT18}
Huan Fu, Mingming Gong, Chaohui Wang, Kayhan Batmanghelich, and Dacheng Tao.
\newblock {Deep Ordinal Regression Network for Monocular Depth Estimation}.
\newblock In {\em {CVPR}}, pages 2002--2011, 2018.

\bibitem{DBLP:conf/mm/GaoZZT21}
Li~Gao, Jing Zhang, Lefei Zhang, and Dacheng Tao.
\newblock {{DSP:} Dual Soft-Paste for Unsupervised Domain Adaptive Semantic
  Segmentation}.
\newblock In {\em {ACM} Multimedia}, pages 2825--2833, 2021.

\bibitem{DBLP:conf/cvpr/GuoYLY21}
Xiaoqing Guo, Chen Yang, Baopu Li, and Yixuan Yuan.
\newblock {MetaCorrection: Domain-Aware Meta Loss Correction for Unsupervised
  Domain Adaptation in Semantic Segmentation}.
\newblock In {\em {CVPR}}, pages 3927--3936, 2021.

\bibitem{DBLP:conf/ijcnn/MarsdenBDY22}
Robert~A. Marsden, Alexander Bartler, Mario D{\"{o}}bler, and Bin Yang.
\newblock {Contrastive Learning and Self-Training for Unsupervised Domain
  Adaptation in Semantic Segmentation}.
\newblock In {\em {IJCNN}}, pages 1--8, 2022.

\bibitem{DBLP:journals/tip/ZhengY22}
Zhedong Zheng and Yi~Yang.
\newblock {Adaptive Boosting for Domain Adaptation: Toward Robust Predictions
  in Scene Segmentation}.
\newblock {\em {IEEE} Trans. Image Process.}, 31:5371--5382, 2022.

\bibitem{DBLP:conf/iccv/0003DSX19}
Hao Lu, Yutong Dai, Chunhua Shen, and Songcen Xu.
\newblock {Indices Matter: Learning to Index for Deep Image Matting}.
\newblock In {\em {ICCV}}, pages 3265--3274, 2019.

\bibitem{DBLP:conf/cvpr/YuKF17}
Fisher Yu, Vladlen Koltun, and Thomas~A. Funkhouser.
\newblock {Dilated Residual Networks}.
\newblock In {\em {CVPR}}, pages 636--644, 2017.

\bibitem{DBLP:journals/corr/YuK15}
Fisher Yu and Vladlen Koltun.
\newblock {Multi-Scale Context Aggregation by Dilated Convolutions}.
\newblock In {\em {ICLR}}, 2016.

\bibitem{DBLP:conf/nips/LuoLUZ16}
Wenjie Luo, Yujia Li, Raquel Urtasun, and Richard~S. Zemel.
\newblock {Understanding the Effective Receptive Field in Deep Convolutional
  Neural Networks}.
\newblock In {\em {NIPS}}, pages 4898--4906, 2016.

\bibitem{DBLP:journals/prl/KimCJLJK23}
Bum~Jun Kim, Hyeyeon Choi, Hyeonah Jang, Dong~Gu Lee, Wonseok Jeong, and
  Sang~Woo Kim.
\newblock {Dead pixel test using effective receptive field}.
\newblock {\em Pattern Recognit. Lett.}, 167:149--156, 2023.

\bibitem{DBLP:journals/corr/SimonyanVZ13}
Karen Simonyan, Andrea Vedaldi, and Andrew Zisserman.
\newblock {Deep Inside Convolutional Networks: Visualising Image Classification
  Models and Saliency Maps}.
\newblock In {\em {ICLR} (Workshop)}, 2014.

\bibitem{DBLP:journals/ijcv/SelvarajuCDVPB20}
Ramprasaath~R. Selvaraju, Michael Cogswell, Abhishek Das, Ramakrishna Vedantam,
  Devi Parikh, and Dhruv Batra.
\newblock {Grad-CAM: Visual Explanations from Deep Networks via Gradient-Based
  Localization}.
\newblock {\em Int. J. Comput. Vis.}, 128(2):336--359, 2020.

\bibitem{ssp}
Yassine.
\newblock Semantic segmentation {in PyTorch}.
\newblock \url{https://github.com/yassouali/pytorch-segmentation}, 2020.

\bibitem{DBLP:conf/nips/PaszkeGMLBCKLGA19}
Adam Paszke, Sam Gross, Francisco Massa, Adam Lerer, James Bradbury, Gregory
  Chanan, Trevor Killeen, Zeming Lin, Natalia Gimelshein, Luca Antiga, Alban
  Desmaison, Andreas K{\"{o}}pf, Edward~Z. Yang, Zachary DeVito, Martin Raison,
  Alykhan Tejani, Sasank Chilamkurthy, Benoit Steiner, Lu~Fang, Junjie Bai, and
  Soumith Chintala.
\newblock {PyTorch: An Imperative Style, High-Performance Deep Learning
  Library}.
\newblock In {\em NeurIPS}, pages 8024--8035, 2019.

\bibitem{DBLP:journals/corr/ChenPSA17}
Liang{-}Chieh Chen, George Papandreou, Florian Schroff, and Hartwig Adam.
\newblock {Rethinking Atrous Convolution for Semantic Image Segmentation}.
\newblock {\em CoRR}, abs/1706.05587, 2017.

\bibitem{DBLP:conf/eccv/ChenZPSA18}
Liang{-}Chieh Chen, Yukun Zhu, George Papandreou, Florian Schroff, and Hartwig
  Adam.
\newblock {Encoder-Decoder with Atrous Separable Convolution for Semantic Image
  Segmentation}.
\newblock In {\em {ECCV} {(7)}}, volume 11211, pages 833--851, 2018.

\bibitem{DBLP:journals/ijcv/EveringhamEGWWZ15}
Mark Everingham, S.~M.~Ali Eslami, Luc~Van Gool, Christopher K.~I. Williams,
  John~M. Winn, and Andrew Zisserman.
\newblock {The Pascal Visual Object Classes Challenge: {A} Retrospective}.
\newblock {\em Int. J. Comput. Vis.}, 111(1):98--136, 2015.

\bibitem{DBLP:conf/nips/ChenCZPZSAS18}
Liang{-}Chieh Chen, Maxwell~D. Collins, Yukun Zhu, George Papandreou, Barret
  Zoph, Florian Schroff, Hartwig Adam, and Jonathon Shlens.
\newblock {Searching for Efficient Multi-Scale Architectures for Dense Image
  Prediction}.
\newblock In {\em NeurIPS}, pages 8713--8724, 2018.

\bibitem{DBLP:conf/cvpr/LiuCSAHY019}
Chenxi Liu, Liang{-}Chieh Chen, Florian Schroff, Hartwig Adam, Wei Hua, Alan~L.
  Yuille, and Li~Fei{-}Fei.
\newblock {Auto-DeepLab: Hierarchical Neural Architecture Search for Semantic
  Image Segmentation}.
\newblock In {\em {CVPR}}, pages 82--92, 2019.

\bibitem{DBLP:conf/cvpr/CordtsORREBFRS16}
Marius Cordts, Mohamed Omran, Sebastian Ramos, Timo Rehfeld, Markus Enzweiler,
  Rodrigo Benenson, Uwe Franke, Stefan Roth, and Bernt Schiele.
\newblock {The Cityscapes Dataset for Semantic Urban Scene Understanding}.
\newblock In {\em {CVPR}}, pages 3213--3223, 2016.

\bibitem{DBLP:conf/cvpr/HeZRS16}
Kaiming He, Xiangyu Zhang, Shaoqing Ren, and Jian Sun.
\newblock {Deep Residual Learning for Image Recognition}.
\newblock In {\em {CVPR}}, pages 770--778, 2016.

\bibitem{DBLP:journals/ijcv/ZhouZPXFBT19}
Bolei Zhou, Hang Zhao, Xavier Puig, Tete Xiao, Sanja Fidler, Adela Barriuso,
  and Antonio Torralba.
\newblock {Semantic Understanding of Scenes Through the {ADE20K} Dataset}.
\newblock {\em Int. J. Comput. Vis.}, 127(3):302--321, 2019.

\bibitem{DBLP:journals/corr/abs-1903-11816}
Huikai Wu, Junge Zhang, Kaiqi Huang, Kongming Liang, and Yizhou Yu.
\newblock {FastFCN: Rethinking Dilated Convolution in the Backbone for Semantic
  Segmentation}.
\newblock {\em CoRR}, abs/1903.11816, 2019.

\bibitem{DBLP:conf/cvpr/ZhaoSQWJ17}
Hengshuang Zhao, Jianping Shi, Xiaojuan Qi, Xiaogang Wang, and Jiaya Jia.
\newblock {Pyramid Scene Parsing Network}.
\newblock In {\em {CVPR}}, pages 6230--6239, 2017.

\bibitem{DBLP:conf/cvpr/YangYZLY18}
Maoke Yang, Kun Yu, Chi Zhang, Zhiwei Li, and Kuiyuan Yang.
\newblock {DenseASPP for Semantic Segmentation in Street Scenes}.
\newblock In {\em {CVPR}}, pages 3684--3692, 2018.

\bibitem{DBLP:conf/cvpr/CaesarUF18}
Holger Caesar, Jasper R.~R. Uijlings, and Vittorio Ferrari.
\newblock {COCO-Stuff: Thing and Stuff Classes in Context}.
\newblock In {\em {CVPR}}, pages 1209--1218, 2018.

\bibitem{DBLP:conf/nips/WangZMLZ21}
Junjue Wang, Zhuo Zheng, Ailong Ma, Xiaoyan Lu, and Yanfei Zhong.
\newblock {LoveDA: {A} Remote Sensing Land-Cover Dataset for Domain Adaptive
  Semantic Segmentation}.
\newblock In {\em NeurIPS Datasets and Benchmarks}, 2021.

\bibitem{DBLP:journals/mia/OrlandoFBKBDFHK20}
Jos{\'{e}}~Ignacio Orlando, Huazhu Fu, Jo{\~{a}}o~Barbosa Breda, Karel van
  Keer, Deepti~R. Bathula, Andr{\'{e}}s Diaz{-}Pinto, Ruogu Fang, Pheng{-}Ann
  Heng, Jeyoung Kim, Joonho Lee, Joonseok Lee, Xiaoxiao Li, Peng Liu, Shuai Lu,
  Balamurali Murugesan, Valery Naranjo, Sai Samarth~R. Phaye, Sharath~M.
  Shankaranarayana, and Hrvoje Bogunovic.
\newblock {{REFUGE} Challenge: {A} unified framework for evaluating automated
  methods for glaucoma assessment from fundus photographs}.
\newblock {\em Medical Image Anal.}, 59, 2020.

\bibitem{DBLP:conf/eccv/YinYCLZLH20}
Minghao Yin, Zhuliang Yao, Yue Cao, Xiu Li, Zheng Zhang, Stephen Lin, and Han
  Hu.
\newblock {Disentangled Non-local Neural Networks}.
\newblock In {\em {ECCV} {(15)}}, volume 12360, pages 191--207, 2020.

\bibitem{DBLP:journals/pami/HuangWWHSLH23}
Zilong Huang, Xinggang Wang, Yunchao Wei, Lichao Huang, Humphrey Shi, Wenyu
  Liu, and Thomas~S. Huang.
\newblock {CCNet: Criss-Cross Attention for Semantic Segmentation}.
\newblock {\em {IEEE} Trans. Pattern Anal. Mach. Intell.}, 45(6):6896--6908,
  2023.

\bibitem{DBLP:journals/corr/abs-1907-12273}
Lang Huang, Yuhui Yuan, Jianyuan Guo, Chao Zhang, Xilin Chen, and Jingdong
  Wang.
\newblock {Interlaced Sparse Self-Attention for Semantic Segmentation}.
\newblock {\em CoRR}, abs/1907.12273, 2019.

\bibitem{DBLP:conf/iccv/ZhuXBHB19}
Zhen Zhu, Mengdu Xu, Song Bai, Tengteng Huang, and Xiang Bai.
\newblock {Asymmetric Non-Local Neural Networks for Semantic Segmentation}.
\newblock In {\em {ICCV}}, pages 593--602, 2019.

\bibitem{DBLP:journals/tmi/HooverKG00}
Adam~W. Hoover, Valentina Kouznetsova, and Michael~H. Goldbaum.
\newblock {Locating Blood Vessels in Retinal Images by Piece-wise Threshold
  Probing of a Matched Filter Response}.
\newblock {\em {IEEE} Trans. Medical Imaging}, 19(3):203--210, 2000.

\bibitem{DBLP:journals/tmi/HooverG03}
Adam~W. Hoover and Michael~H. Goldbaum.
\newblock {Locating the Optical Nerve in a Retinal Image Using the Fuzzy
  Convergence of the Blood Vessels}.
\newblock {\em {IEEE} Trans. Medical Imaging}, 22(8):951--958, 2003.

\bibitem{DBLP:journals/tbe/YanYC18}
Zengqiang Yan, Xin Yang, and Kwang{-}Ting Cheng.
\newblock {Joint Segment-Level and Pixel-Wise Losses for Deep Learning Based
  Retinal Vessel Segmentation}.
\newblock {\em {IEEE} Trans. Biomed. Eng.}, 65(9):1912--1923, 2018.

\bibitem{DBLP:conf/miccai/RonnebergerFB15}
Olaf Ronneberger, Philipp Fischer, and Thomas Brox.
\newblock {U-Net: Convolutional Networks for Biomedical Image Segmentation}.
\newblock In {\em {MICCAI} {(3)}}, volume 9351, pages 234--241, 2015.

\bibitem{mmseg2020}
MMSegmentation Contributors.
\newblock {MMSegmentation}: Openmmlab {Semantic Segmentation Toolbox and
  Benchmark}.
\newblock \url{https://github.com/open-mmlab/mmsegmentation}, 2020.

\bibitem{DBLP:journals/tbe/FrazRHUROB12}
Muhammad~Moazam Fraz, Paolo Remagnino, Andreas Hoppe, Bunyarit Uyyanonvara,
  Alicja~R. Rudnicka, Christopher~G. Owen, and Sarah Barman.
\newblock {An Ensemble Classification-Based Approach Applied to Retinal Blood
  Vessel Segmentation}.
\newblock {\em {IEEE} Trans. Biomed. Eng.}, 59(9):2538--2548, 2012.

\bibitem{DBLP:journals/ijbi/BudaiBMHM13}
Attila Budai, R{\"{u}}diger Bock, Andreas~K. Maier, Joachim Hornegger, and
  Georg Michelson.
\newblock {Robust Vessel Segmentation in Fundus Images}.
\newblock {\em Int. J. Biomed. Imaging}, 2013:154860:1--154860:11, 2013.

\bibitem{DBLP:conf/midl/BreaJ0W20}
Luisa~S{\'{a}}nchez Brea, Danilo Andrade~De Jesus, Stefan Klein, and Theo van
  Walsum.
\newblock {Deep learning-based retinal vessel segmentation with cross-modal
  evaluation}.
\newblock In {\em {MIDL}}, volume 121, pages 709--720, 2020.

\bibitem{DBLP:conf/cvpr/DengDSLL009}
Jia Deng, Wei Dong, Richard Socher, Li{-}Jia Li, Kai Li, and Li~Fei{-}Fei.
\newblock {ImageNet: {A} large-scale hierarchical image database}.
\newblock In {\em {CVPR}}, pages 248--255, 2009.

\bibitem{DBLP:conf/cvpr/0005DSZWTA18}
Hang Zhang, Kristin~J. Dana, Jianping Shi, Zhongyue Zhang, Xiaogang Wang,
  Ambrish Tyagi, and Amit Agrawal.
\newblock {Context Encoding for Semantic Segmentation}.
\newblock In {\em {CVPR}}, pages 7151--7160, 2018.

\bibitem{DBLP:conf/cvpr/ZamirAGKSK00XB19}
Syed~Waqas Zamir, Aditya Arora, Akshita Gupta, Salman~H. Khan, Guolei Sun,
  Fahad~Shahbaz Khan, Fan Zhu, Ling Shao, Gui{-}Song Xia, and Xiang Bai.
\newblock {iSAID: {A} Large-scale Dataset for Instance Segmentation in Aerial
  Images}.
\newblock In {\em {CVPR} Workshops}, pages 28--37, 2019.

\bibitem{DBLP:conf/cvpr/XiaBDZBLDPZ18}
Gui{-}Song Xia, Xiang Bai, Jian Ding, Zhen Zhu, Serge~J. Belongie, Jiebo Luo,
  Mihai Datcu, Marcello Pelillo, and Liangpei Zhang.
\newblock {{DOTA:} {A} Large-Scale Dataset for Object Detection in Aerial
  Images}.
\newblock In {\em {CVPR}}, pages 3974--3983, 2018.

\bibitem{DBLP:conf/nips/GuoLHLC022}
Meng{-}Hao Guo, Cheng{-}Ze Lu, Qibin Hou, Zhengning Liu, Ming{-}Ming Cheng, and
  Shi{-}Min Hu.
\newblock {SegNeXt: Rethinking Convolutional Attention Design for Semantic
  Segmentation}.
\newblock In {\em NeurIPS}, 2022.

\bibitem{DBLP:journals/pami/ShelhamerLD17}
Evan Shelhamer, Jonathan Long, and Trevor Darrell.
\newblock {Fully Convolutional Networks for Semantic Segmentation}.
\newblock {\em {IEEE} Trans. Pattern Anal. Mach. Intell.}, 39(4):640--651,
  2017.

\bibitem{DBLP:conf/cvpr/0004GGH18}
Xiaolong Wang, Ross~B. Girshick, Abhinav Gupta, and Kaiming He.
\newblock {Non-Local Neural Networks}.
\newblock In {\em {CVPR}}, pages 7794--7803, 2018.

\bibitem{DBLP:conf/cvpr/FuLT0BFL19}
Jun Fu, Jing Liu, Haijie Tian, Yong Li, Yongjun Bao, Zhiwei Fang, and Hanqing
  Lu.
\newblock {Dual Attention Network for Scene Segmentation}.
\newblock In {\em {CVPR}}, pages 3146--3154, 2019.

\bibitem{newville2016lmfit}
Matthew Newville, Till Stensitzki, Daniel~B Allen, Michal Rawlik, Antonino
  Ingargiola, and Andrew Nelson.
\newblock {LMFIT: Non-linear {least-square minimization and curve-fitting for
  Python}}.
\newblock {\em Astrophysics Source Code Library}, pages ascl--1606, 2016.

\bibitem{DBLP:conf/eccv/XiaoLZJS18}
Tete Xiao, Yingcheng Liu, Bolei Zhou, Yuning Jiang, and Jian Sun.
\newblock {Unified Perceptual Parsing for Scene Understanding}.
\newblock In {\em {ECCV} {(5)}}, volume 11209, pages 432--448, 2018.

\bibitem{DBLP:conf/iclr/DosovitskiyB0WZ21}
Alexey Dosovitskiy, Lucas Beyer, Alexander Kolesnikov, Dirk Weissenborn,
  Xiaohua Zhai, Thomas Unterthiner, Mostafa Dehghani, Matthias Minderer, Georg
  Heigold, Sylvain Gelly, Jakob Uszkoreit, and Neil Houlsby.
\newblock {An Image is Worth 16x16 Words: Transformers for Image Recognition at
  Scale}.
\newblock In {\em {ICLR}}, 2021.

\bibitem{DBLP:conf/icml/TouvronCDMSJ21}
Hugo Touvron, Matthieu Cord, Matthijs Douze, Francisco Massa, Alexandre
  Sablayrolles, and Herv{\'{e}} J{\'{e}}gou.
\newblock {Training data-efficient image transformers {\&} distillation through
  attention}.
\newblock In {\em {ICML}}, volume 139, pages 10347--10357, 2021.

\bibitem{DBLP:conf/iccv/LiuL00W0LG21}
Ze~Liu, Yutong Lin, Yue Cao, Han Hu, Yixuan Wei, Zheng Zhang, Stephen Lin, and
  Baining Guo.
\newblock {Swin Transformer: Hierarchical Vision Transformer using Shifted
  Windows}.
\newblock In {\em {ICCV}}, pages 9992--10002, 2021.

\bibitem{DBLP:conf/cvpr/ZhengLZZLWFFXT021}
Sixiao Zheng, Jiachen Lu, Hengshuang Zhao, Xiatian Zhu, Zekun Luo, Yabiao Wang,
  Yanwei Fu, Jianfeng Feng, Tao Xiang, Philip H.~S. Torr, and Li~Zhang.
\newblock {Rethinking Semantic Segmentation From a Sequence-to-Sequence
  Perspective With Transformers}.
\newblock In {\em {CVPR}}, pages 6881--6890, 2021.

\bibitem{DBLP:conf/iccv/StrudelPLS21}
Robin Strudel, Ricardo~Garcia Pinel, Ivan Laptev, and Cordelia Schmid.
\newblock {Segmenter: Transformer for Semantic Segmentation}.
\newblock In {\em {ICCV}}, pages 7242--7252, 2021.

\bibitem{DBLP:conf/iccv/RanftlBK21}
Ren{\'{e}} Ranftl, Alexey Bochkovskiy, and Vladlen Koltun.
\newblock {Vision Transformers for Dense Prediction}.
\newblock In {\em {ICCV}}, pages 12159--12168, 2021.

\bibitem{DBLP:conf/nips/XieWYAAL21}
Enze Xie, Wenhai Wang, Zhiding Yu, Anima Anandkumar, Jose~M. Alvarez, and Ping
  Luo.
\newblock {SegFormer: Simple and Efficient Design for Semantic Segmentation
  with Transformers}.
\newblock In {\em NeurIPS}, pages 12077--12090, 2021.

\end{thebibliography}
\end{document}